\documentclass[10pt,twocolumn,letterpaper]{article}

\usepackage{cvpr}              

\usepackage[accsupp]{axessibility}  
\definecolor{cvprblue}{rgb}{0.21,0.49,0.74}
\usepackage[pagebackref,breaklinks,colorlinks,allcolors=cvprblue]{hyperref}

\usepackage{algpseudocode}
\usepackage{algorithm}
\usepackage{makecell}
\def\paperID{11945} 
\def\confName{CVPR}
\def\confYear{2025}

\title{3D-Mem: 3D Scene Memory for Embodied Exploration and Reasoning}

\author{
\textbf{Yuncong Yang}$^{1}$\thanks{\:\:Equal Contribution}, \textbf{Han Yang}$^{2}$\footnotemark[1], \textbf{Jiachen Zhou}$^{3}$, \textbf{Peihao Chen}$^{1}$,  \\\textbf{Hongxin Zhang}$^{1}$, 
    \textbf{Yilun Du}$^{4}$, \textbf{Chuang Gan}$^{1, 5}$ \\
    $^1$UMass Amherst, $^2$CUHK, $^3$Columbia University \\
    $^4$MIT, $^5$MIT-IBM Watson AI Lab\\
    \texttt{yuncongyang@umass.edu}
}

\begin{document}
\twocolumn[{
\renewcommand\twocolumn[1][]{#1}
\maketitle
\begin{center}

\vspace{-1cm}
\href{https://umass-embodied-agi.github.io/3D-Mem/}{\large https://umass-embodied-agi.github.io/3D-Mem/}
\vspace{7mm}

\includegraphics[width=0.80\textwidth]{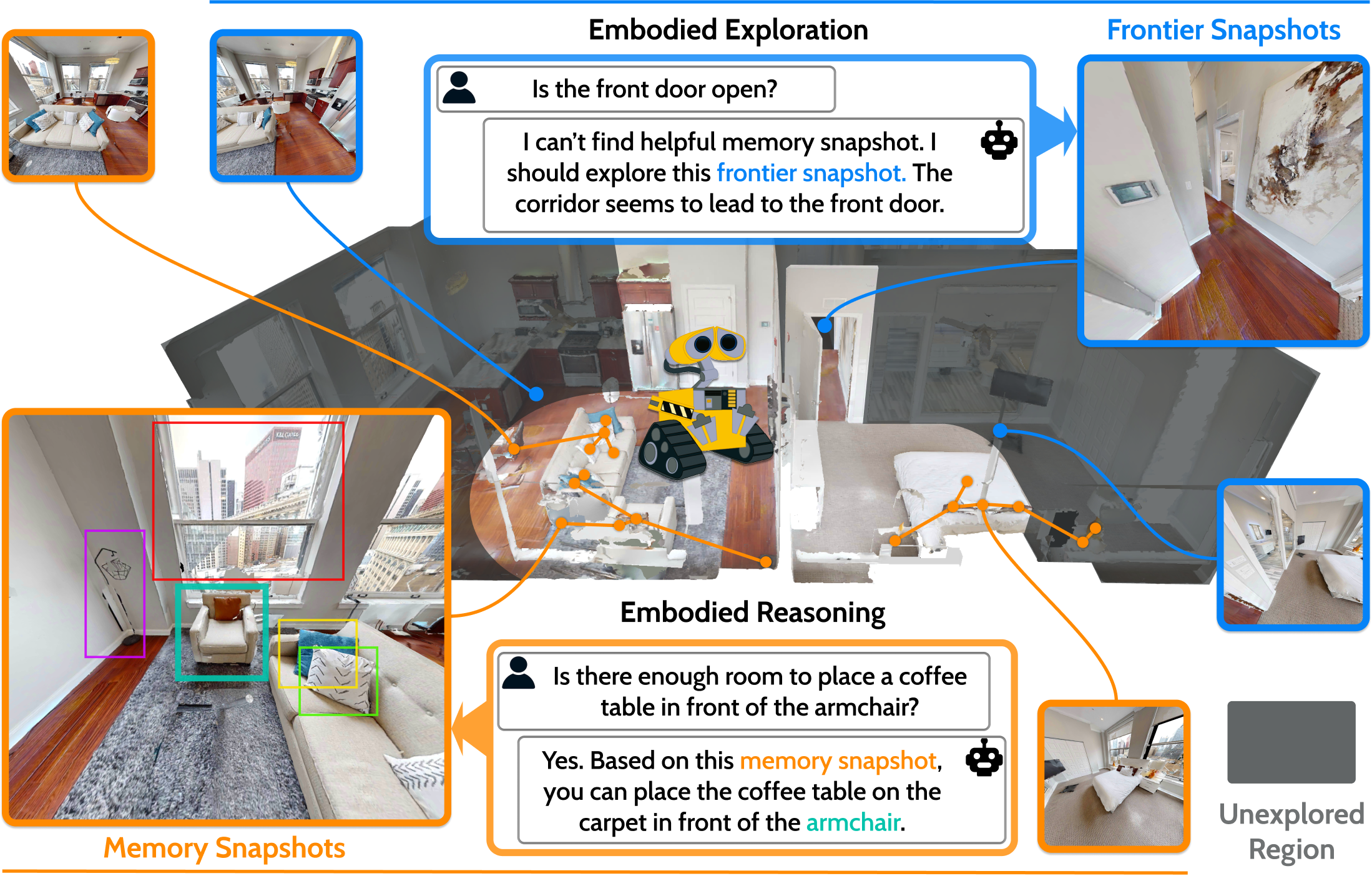}
\captionof{figure}{With 3D-Mem, explored regions are represented by a set of Memory Snapshots capturing clusters of co-visible objects, \textit{i.e.}, the objects observable in a single image observation, along with their spatial relationships and background context, as shown in the bottom-left example. Unexplored regions are represented by navigable frontiers along with image observations, referred to as Frontier Snapshots.} 
\label{fig:teaser}
\end{center}
}]

\renewcommand{\thefootnote}{*}
\footnotetext{Equal Contribution}

\renewcommand{\thefootnote}{\arabic{footnote}}

\begin{abstract}
Constructing compact and informative 3D scene representations is essential for effective embodied exploration and reasoning, especially in complex environments over extended periods. Existing representations, such as object-centric 3D scene graphs, oversimplify spatial relationships by modeling scenes as isolated objects with restrictive textual relationships, making it difficult to address queries requiring nuanced spatial understanding. Moreover, these representations lack natural mechanisms for active exploration and memory management, hindering their application to lifelong autonomy. In this work, we propose 3D-Mem, a novel 3D scene memory framework for embodied agents. 3D-Mem employs informative multi-view images, termed Memory Snapshots, to capture rich visual information of explored regions. It further integrates frontier-based exploration by introducing Frontier Snapshots—glimpses of unexplored areas—enabling agents to make decisions by considering both known and potential new information. To support lifelong memory in active exploration settings, we present an incremental construction pipeline for 3D-Mem, as well as a memory retrieval technique for memory management. Experimental results on three benchmarks demonstrate that 3D-Mem significantly enhances agents' exploration and reasoning capabilities in 3D environments, highlighting its potential for advancing applications in embodied AI.
\end{abstract}    
\section{Introduction}
\label{sec:intro}
Embodied agents operating in complex 3D environments require robust and lifelong scene memory to store rich visual information about the environment, supporting effective exploration and reasoning over extended periods. 
Currently, there are two main streams of scene representations. The first stream focuses on object-centric representations, particularly 3D scene graphs \citep{wald2020learning, conceptgraph}, that represent scenes using nodes for objects and edges for inter-object relationships. Another stream directly uses dense 3D representations, such as point clouds \citep{ding2023pla,zhang2023clip,ding2024lowis3d,jatavallabhula2023conceptfusion} or neural fields \citep{tsagkas2023vl, kerr2023lerf,mazur2023feature}, filling 3D space with dense visual features for querying. 


However, these scene representations exhibit significant limitations when employed as scene memory for embodied agents. Object-centric representations like 3D scene graphs tend to oversimplify 3D scenes by modeling them as individual objects and quantizing inter-object relationships into restrictive textual descriptions. This approach results in the loss of critical spatial information, making it difficult to answer questions that require an understanding of spatial relationships. As illustrated in the bottom-left example from Figure~\ref{fig:teaser}, when attempting to answer the question \textit{``Is there enough room to place a coffee table in front of the armchair?"} using a 3D scene graph, the only spatial-related information we can access are the numerical 3D bounding boxes of the objects and the textual descriptions of the spatial relationships among them. It is challenging to measure unoccupied space or determine oriented spatial relationships such as \textit{``in front of"} based solely on this information. On the other hand, dense 3D representations like point clouds or neural fields \citep{yang2024llm, xu2025pointllm, 3dllm, huang2023embodied} are computationally expensive and lack scalability as the scene grows during the agent's exploration. Meanwhile, unlike vision-language models (VLMs) and large language models (LLMs) that effectively reason over images and text, current foundation models lack sufficient reasoning capabilities for dense 3D modalities due to limited training data. Additionally, neither type of existing scene representation can model unexplored regions, thus failing to support agents in active exploration, which restrains agents from leveraging their scene memory to expand their knowledge and achieve their goals for embodied tasks. 


To address these challenges, we introduce \textbf{3D-Mem}, a more capable 3D scene memory for embodied agents. 3D-Mem employs a set of informative multi-view images, termed \textbf{Memory Snapshots}, to encompass the explored regions of a 3D scene. Each memory snapshot captures all objects visible in that snapshot along with their surroundings. Our intuition is that a snapshot image alone is sufficient to capture rich visual information of a region. 
For example, revisiting the challenge illustrated in Figure~\ref{fig:teaser}, a memory snapshot clearly shows that there is sufficient space \textit{``in front of"} the armchair to place a coffee table. With the recent advancements in VLMs, these models can extract such spatial information from images \citep{chen2024spatialvlm, OpenEQA2023}, much like humans do through intuitive observations.  



Building upon the concept of employing multi-view images as 3D scene memory, 3D-Mem integrates with frontier-based exploration frameworks and extends the concept of ``frontiers" to \textbf{Frontier Snapshots}, which represent glimpses of unexplored regions in a 3D scene. In the scenario depicted in the top-left example from Figure~\ref{fig:teaser}, we can observe that the frontier snapshot corresponds to a corridor, which is the most probable path leading to the front door. By maintaining these frontier snapshots, the agent can choose either to complete tasks based on its accumulated knowledge or navigate to unexplored regions for new information, as illustrated in the embodied question-answering tasks in Figure~\ref{fig:teaser}. Meanwhile, representing both explored and unexplored regions through multi-view images allows us to better leverage the decision-making capability of VLMs, enhancing the agent's ability to reason and plan effectively.

Additionally, as agents in lifelong settings continuously explore and expand their knowledge, the 3D scene memory system must operate effectively as the memory grows. 
To achieve this, 3D-Mem supports real-time incremental memory aggregation. To efficiently manage the ever-expanding scene memory, we introduce Prefiltering as an effective memory retrieval mechanism to select relevant memory during decision-making.
This framework enables the agent to perform continuous exploration and navigation over extended periods without excessive computational burdens.

Extensive experiments and superior performance on three benchmarks demonstrate that 3D-Mem significantly enhances agents' capabilities in reasoning and lifelong exploration within 3D environments. 

Our contributions can be summarized as follows:
\begin{itemize}[align=right,itemindent=0em,labelsep=3pt,labelwidth=0em,leftmargin=1em,itemsep=0em] 
    \item We introduce 3D-Mem, a compact scene memory that constructs informative multi-view snapshot images to capture diverse and robust information among co-visible objects and their surroundings in 3D scenes.
    \item By introducing frontier snapshots to include unexplored regions, 3D-Mem enables agents to actively explore and acquire new information. 
    \item We incorporate 3D-Mem with incremental memory aggregation and prefiltering strategies that enable agents to expand their knowledge and adapt over extended periods, supporting lifelong learning in 3D environments.
\end{itemize}
\section{Related Work}
\label{sec:rel_works}
\noindent \textbf{3D Scene Representations.} 
Recent works \citep{peng2023openscene, zhang2023clip, shafiullah2022clip, yamazaki2024open} have focused on establishing dense 3D representations by grounding 2D representations captured by Vision-Language Foundation Models (\eg CLIP~\cite{radford2021learning}, BLIP~\cite{li2022blip}, SEEM~\cite{zou2024segment}) into 3D scenes, which showcase impressive results on tasks such as open-vocabulary object segmentation and language-guided object grounding \citep{hong20223dgrounding}. However, such representations are limited due to high resource consumption and the inability to support dynamic updates. 3D scene graphs address these limitations by formulating the scene as a compact graph, where nodes represent objects, and edges encode inter-object relationships as textual descriptions \citep{gay2019visual, armeni20193d, kim20193, wald2020learning, conceptgraph}, enabling real-time establishment and dynamic update for hierarchical scene representations \citep{rosinol2021kimera,wu2021scenegraphfusion,hughes2022hydra}. While such object-centric representations have demonstrated effectiveness in various tasks, they remain constrained for oversimplifying inter-object relationships with restrictive text descriptions. To tackle this challenge, our work leverages a set of informative memory snapshots to visually capture spatial and semantic relationships among objects, offering a more sophisticated understanding of the scene.

\noindent \textbf{VLM for Exploration and Reasoning.}
 Vision-Language Models (VLMs) have shown promising results in solving embodied exploration and reasoning tasks by leveraging commonsense reasoning and internet-scale knowledge. Existing exploration approaches can be divided into two categories. The former directly employs consecutive observations together with instructions as input, requiring the VLM to predict the next-step action \citep{zhang2024navid} while the latter grounds the exploration target in the 3D scene through visual prompting, establishing a semantic map to guide the exploration process \citep{majumdar2022zson,shah2023navigation,ren2024explore, yokoyama2024vlfm}. However, both approaches are constrained by their memory representations. For the former, vanilla past observations can only serve as short-term memory. For the latter, their semantic maps are target-specific and cannot be generalized to future tasks. 
 To address these limitations, our work introduces the first lifelong and target-agnostic scene memory that can be seamlessly integrated with VLM for further reasoning, stepping closer to the ultimate goal of lifelong autonomy.

\noindent \textbf{Topological Mapping.} Besides the 3D scene representations mentioned above, prior 2D topological mapping methods like TSGM~\cite{TSGM} and RoboHop ~\cite{RoboHop} provide valuable context. They construct navigation graphs from images and objects but do not fully capture all objects or their interrelationships. In contrast, 3D-Mem clusters co-visible objects to represent inter-object spatial relationships, enabling tasks such as embodied Q\&A that extend beyond route planning. Moreover, its efficient memory retrieval mechanism scales to lifelong exploration, leveraging VLMs for advanced reasoning and decision-making.
\section{Approach}
\label{sec:approach}

3D-Mem contains two types of snapshots: \textbf{Memory Snapshots} and \textbf{Frontier Snapshots}. A memory snapshot represents a cluster of objects and their surroundings in the explored regions, while a frontier snapshot represents an unexplored region along with its exploration direction. During exploration, we construct the set of memory snapshots from a stream of RGB-D images with poses and maintain frontier snapshots using frontier algorithms and occupancy maps. Given an objective and the set of memory and frontier snapshots, a VLM agent iteratively selects the most promising frontier snapshot and moves toward it, continually updating both sets of snapshots while making decisions, until the objective is achieved based on the current memory snapshots.

First, in Section~\ref{approach_mem_construction}, we introduce how the set of memory snapshots is constructed from a series of RGB-D observations with
poses. Next, in Section~\ref{approach_integrate_with_explore}, we extend this algorithm to incremental construction in active exploration settings. By integrating frontier snapshots, we enable the agent to actively conduct frontier-based exploration. We also design methods to efficiently retrieve memory as it scales up. Finally, in Section~\ref{approach_explore_reasoning}, we detail how the agent utilizes the constructed memory and frontier snapshots for exploration and reasoning.



\begin{algorithm}
\caption{Co-Visibility Clustering for Memory Snapshots}\label{pseudo:static}
\begin{algorithmic}[1]
    \small
    \State Initial clusters $\mathcal{C} = \{\mathcal{O}\}$
    \State Memory snapshot set $\mathcal{S} = \varnothing$
    \State All frame candidates $\mathcal{I}$
    \State Score function $\mathcal{F}$
    \While{$\mathcal{C}$ is not empty}
        \State $\mathcal{O}^* = \arg\max_{\mathcal{O} \in \mathcal{C}} |\mathcal{O}|$
        \State $\mathcal{I}^* = \{I| I \in \mathcal{I}, \mathcal{O}^* \subseteq \mathcal{O}_{I} \}$
        \If{$\mathcal{I}^*$ is not empty}
            \State $I^* = \arg\max_{I \in \mathcal{I}^*} \mathcal{F}(I)$
            \State $S^* = \langle \mathcal{O}^*, I^*\rangle$
            \State $\mathcal{S} = \mathcal{S} \cup \{S^*\}$
        \Else 
            \State Use K-Means to split $\mathcal{O}^*$ into two clusters $\mathcal{O}^* = \mathcal{O}_1^* \cup \mathcal{O}_2^*$ based on 2D horizontal positions $(x,y)$ 
            \State $\mathcal{C} = \mathcal{C} \cup \{\mathcal{O}_1^*, \mathcal{O}_2^*\}$
        \EndIf

        \State $\mathcal{C} = \mathcal{C} - \{\mathcal{O}^*\}$
    \EndWhile
    \State \While{exists $S_j, S_k \in \mathcal{S}$ such that $I_{S_j} = I_{S_k}$}
    \State  $\mathcal{S} = \mathcal{S} \setminus \{S_j, S_k \}$ 
    \State  $S_l = \langle \mathcal{O}_{S_j} \cup \mathcal{O}_{S_k}, I_{S_j} \rangle$ 
    \State  $\mathcal{S} = \mathcal{S} \cup \{S_l\}$

    \EndWhile

\noindent\Return $\mathcal{S}$
\end{algorithmic}
\end{algorithm}

\subsection{3D-Mem Construction}
\label{approach_mem_construction}
Inspired by the idea that an image itself inherently contains rich and robust information to represent a small area of a scene, we propose a novel way that utilizes a set of multi-view snapshot images to cover the whole informative areas of a scene. Instead of the object-centric representation proposed by ConceptGraph\citep{conceptgraph}, in which only object-level visual features are stored and managed, we propose using one snapshot image to represent a cluster of co-visible objects within that image, namely a \textbf{Memory Snapshot}. This approach not only encapsulates foreground object-level details but also captures contextual room-level cues embedded in the background, providing a more holistic representation of the scene.
\subsubsection{Memory Snapshot Formulation}
Specifically, given a set of $N$ image observations $\mathcal{I}^{obs} = \{I_1^{obs}, I_2^{obs}, ..., I_N^{obs}\}$, where each $I_i^{obs}$ consists of an RGB-D image and its pose, we aim to construct a set of memory snapshots, a minimal subset of $\mathcal{I}^{obs}$, to cover all salient observed objects. 
We first follow ConceptGraph~\citep{conceptgraph} to perform a series of object segmentation, spatial transformations, matching, and merging, resulting in an object set that contains all detected objects from the observations $\mathcal{O} =  \{o_1, o_2, ..., o_M\}$, where each object is characterized by an object category, detection confidence, and 3D location. 
Meanwhile, we obtain a set of frame candidates $\mathcal{I} = \{ I_1, I_2, ..., I_N \}$, where each $I_i = \langle I_i^{obs}, \mathcal{O}_{I_i} \rangle$ consists of the image observation $I_i^{obs}$ together with a set of all detected objects $\mathcal{O}_{I_i}$, \textit{i.e.}, all co-visible objects in $I_i^{obs}$. 

We define $\mathcal{S}$ as a set of memory snapshots $\{ S_1, S_2, ..., S_K \}$ of size $K \leq N$, where each memory snapshot $S_k = \langle \mathcal{O}_{S_k}, I_{S_k} \rangle$ is characterized by a frame candidate $I_{S_k} \in \mathcal{I}$ and a cluster of objects $\mathcal{O}_{S_k}$, a subset of all detected objects $\mathcal{O}_{I_{S_k}}$ in the image $I_{S_k}^{obs}$. Therefore, an image $I_{S_k}^{obs}$ serves as a shared visual feature among $\mathcal{O}_{S_k}$. Since $\mathcal{S}$ needs to cover the whole object set $\mathcal{O}$, and each object $o_j$ needs to be uniquely mapped to one memory snapshot $S_k$ (although it may still be visible in other memory snapshots), we require $\mathcal{O}_{S_1} \cup \mathcal{O}_{S_2} \cup ... \cup \mathcal{O}_{S_K} = \mathcal{O}$, and $\mathcal{O}_{S_i} \cap \mathcal{O}_{S_j} = \varnothing$ for $\forall S_i, S_j \in \mathcal{S}$.
\subsubsection{Co-Visibility Clustering}
To acquire the desired set of memory snapshots, we designed Co-Visibility Clustering based on the classical hierarchical clustering \cite{bisectingkmeans} to split $\mathcal{O}$ into clusters, each of which is a subset of the detected object set $\mathcal{O}_{I_i}$ of a certain frame candidate $I_i$. As detailed in the pseudocode in Algorithm~\ref{pseudo:static}, we define a cluster set $\mathcal{C}$ composed of all unsettled object clusters that haven't been matched with observations, initialized to contain the full object set $\{ \mathcal{O}\}$, and the memory snapshot set $\mathcal{S}$, initialized to $\varnothing$. 
Each time, we pick the largest unsettled cluster $\mathcal{O}^*$ from $\mathcal{C}$ and search through all frame candidates for capable candidates $I^*$ such that $\mathcal{O}^*$ is a subset of the detected object list of $I^*$. When such candidates exist, we rank them based on a score function $\mathcal{F}$ and pick the top-ranked frame candidate $I^*$ to create a new memory snapshot $S^* = \langle \mathcal{O}^*, I^* \rangle$ and add it to $\mathcal{S}$. 
In practice, we choose $\mathcal{F}(I_i) = |\mathcal{O}_{I_i}|$ to select the frame candidate that contains the most objects, and when more than one frame candidate has the same largest object number, we choose the one with the highest sum of detection confidence. 
If there is no feasible frame candidate, we then use K-Means to further divide $\mathcal{O}^*$ into two subclusters $\mathcal{O}_1^*$ and $\mathcal{O}_2^*$ based on the 2D horizontal positions of the objects, and add them to $\mathcal{C}$. We repeat the above process until no clusters remain in $\mathcal{C}$. Note that the process is guaranteed to terminate since every object has been captured in certain observations. Ultimately, after all objects have been assigned to corresponding snapshots, memory snapshots sharing the same frame candidates are merged to achieve the final compact memory representation $\mathcal{S}$.


In each memory snapshot, not only is the visual information of each object stored, but also the spatial relationships between objects and the room-level information are provided by visual cues in the background.
With the increasing perception abilities of VLMs, such multi-view images representations can provide richer and more robust visual information for VLMs to complete difficult tasks. 


\subsection{3D-Mem with Frontier-based Exploration}
\label{approach_integrate_with_explore}


We adapt the memory construction algorithm introduced above to the frontier-based exploration pipeline~\cite{ren2024explore}, where we introduce frontier snapshot as an extension of frontier (Section~\ref{approach_frontier_snapshot}).
In the frontier-based exploration episode, an agent is initialized in an unknown scene and explores the environment step by step. At each step, the agent receives a series of observations, including rgb, depth, and pose, which are used to update frontier and memory snapshots (Section~\ref{approach_scene_graph_update}). As the 3D scene memory grows during exploration, the decision-making becomes computationally heavy. Therefore we introduce Prefiltering as an efficient memory retrieval mechanism (Section~\ref{approach_prefiltering}). 

\subsubsection{Introduction of Frontier Snapshot}
\label{approach_frontier_snapshot}

\begin{figure*}
    \centering
    \includegraphics[width=0.8\linewidth]{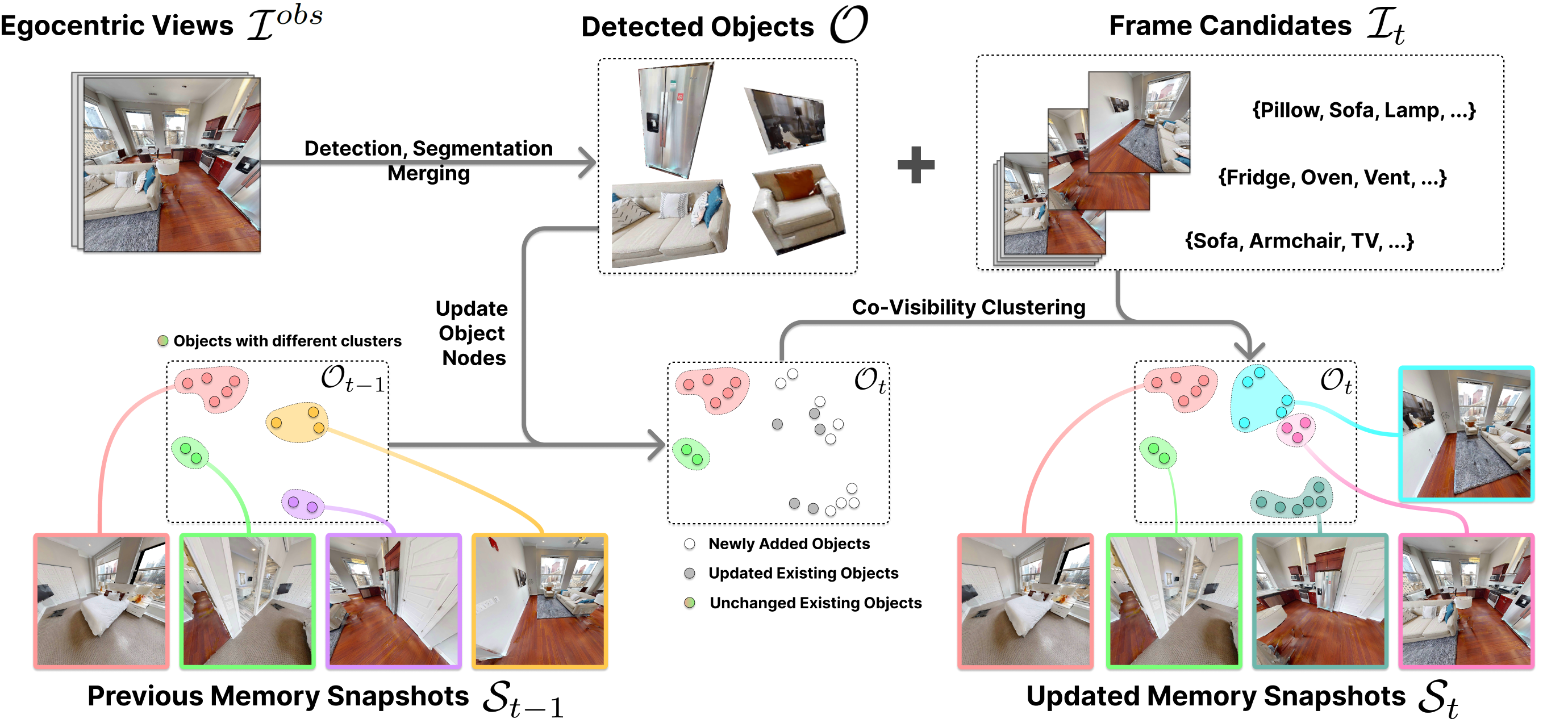}
    \vspace{-10pt}
    \caption{\textbf{The memory aggregation process of 3D-Mem.} At each step $t$, the object set $\mathcal{O}_t$ is first updated using the object-wise update pipeline introduced in Section~\ref{approach_mem_construction}. The newly detected objects and the updated existing objects are then jointly clustered into new memory snapshots using co-visibility clustering (Algorithm~\ref{pseudo:static}), which are used to update the memory snapshot set $\mathcal{S}_t$.}
    \label{fig:update_snapshot}
    \vspace{-15pt}
\end{figure*}

In a common frontier-based exploration framework, an occupancy map is kept to record the explored regions, defined as the nearby areas along the agent's trajectory, and the unexplored regions, defined as navigable but yet-to-be-explored areas. A frontier is then defined to represent such an unexplored region that could be further explored. In this work, we extend this concept by using a snapshot to represent a frontier, similar to memory snapshots. We define a \textbf{Frontier Snapshot} $F = \langle r, p, I^{obs} \rangle $, consisting of the unexplored region $r$ it represents, a navigable location $p$, and an image observation $I^{obs} $ from the agent's position toward that unexplored region. Therefore, the frontier and memory snapshots, which are all raw images, can be directly used as visual input for the VLMs. More details about frontier-based exploration are provided in Appendix~\ref{appendix_frontier_details}.



\subsubsection{Incremental Construction of 3D-Mem}
\label{approach_scene_graph_update}

Throughout the exploration process, the scene memory is dynamically and incrementally constructed. At each exploration step, the agent observes its surroundings and updates the scene memory and frontiers. At step $t$, we denote the current object set as $\mathcal{O}_t$, the frontier set as $\mathcal{F}_t$, the memory snapshot set as $\mathcal{S}_t$, and the frame candidate set as $\mathcal{I}_t$, all of which are initialized as $\varnothing$ at the beginning of the episode.

\noindent\textbf{Object Update.}
As illustrated in Figure~\ref{fig:update_snapshot}, at each time step $t$, the agent first captures $N$ egocentric views $\mathcal{I}^{obs} = \{ I_1^{obs}, I_2^{obs}, ..., I_{N}^{obs} \}$, which are used to extract the object set $\mathcal{O}$ and the frame candidate set $\mathcal{I}$ similar to Section~\ref{approach_mem_construction}. Specifically, a threshold ``$max\_dist$" is set to ensure only objects within a certain distance from the agent are added to the scene graph, as the memory snapshot should only represent objects from a local area. It is important to note that the object set $\mathcal{O}$ detected in these egocentric views may contain both newly identified objects and those already present in the previous set $\mathcal{O}_{t-1}$. Subsequently, the full object set and frame candidate set are updated as $\mathcal{O}_{t} = \mathcal{O}_{t-1} \cup \mathcal{O} $ and $ \mathcal{I}_{t} = \mathcal{I}_{t-1} \cup \mathcal{I} $, respectively.

\noindent\textbf{Memory Snapshot Update.} 
We implement the co-visibility clustering in Section~\ref{approach_mem_construction} incrementally. At each time step $t$, instead of performing clustering on the entire object set $\mathcal{O}_t$, we focus on clustering objects related to $\mathcal{O}$, the objects detected from the egocentric views at this step. 
In $\mathcal{O}$, some objects may have already been assigned to specific memory snapshots in $\mathcal{S}_{t-1}$. We refer to those memory snapshots as $\mathcal{S}_{prev} = \{S | S \in \mathcal{S}_{t-1}, \mathcal{O}_S \cap \mathcal{O} \neq \varnothing \} $.
All objects from $\mathcal{S}_{prev}$, along with the newly detected objects in $\mathcal{O}$, are used as input for clustering, denoted as $\mathcal{O}_{input}$. Then, the memory snapshot set is updated as $\mathcal{S}_t = (\mathcal{S}_{t-1} \setminus \mathcal{S}_{prev}) \cup \text{Cluster}(\mathcal{O}_{input}, \mathcal{I}_t) $

\noindent\textbf{Frontier Snapshot Update.}
At each step $t$, an existing frontier from $\mathcal{F}_{t-1}$ may be modified if the unexplored region it represents has been updated, or it may be removed if the region has been fully explored. Additionally, new frontiers may be introduced. For each newly added or modified frontier, a new snapshot is taken to update its image representation. Then, $\mathcal{F}_{t-1}$ is updated to $\mathcal{F}_t$. More implementation details are in Appendix~\ref{appendix_vlm_navigation}.



\subsubsection{Memory Retrieval with Prefiltering}
\label{approach_prefiltering}
During exploration, when we incrementally construct memory snapshots, the large volumes of memory can make the agent's decision-making inefficient and introduce noise unrelated to the objectives. Therefore, we designed a memory retrieval mechanism called \textbf{Prefiltering}. Since most memory snapshots are irrelevant to a given instruction and processing them consumes substantial computational resources without meaningful benefit, we present the VLM agent with the question along with all object categories in $\mathcal{O}_t$. The VLM agent then outputs the relevant object categories ranked by importance, retaining only the top $K$ categories determined by a hyperparameter $K$. Memory snapshots that do not contain any of these $K$ selected categories are filtered out. Figure~\ref{fig:query_llm} illustrates Prefiltering in an embodied question-answering task, showing how the original set of memory snapshots is reduced based on relevancy. This technique significantly reduces resource consumption, allowing us to include images directly within the prompt and enhancing the agent's efficiency.

\subsection{Exploration and Reasoning with 3D-Mem}
\label{approach_explore_reasoning}

\begin{figure}[ht]
    \centering
    \includegraphics[width=\linewidth]{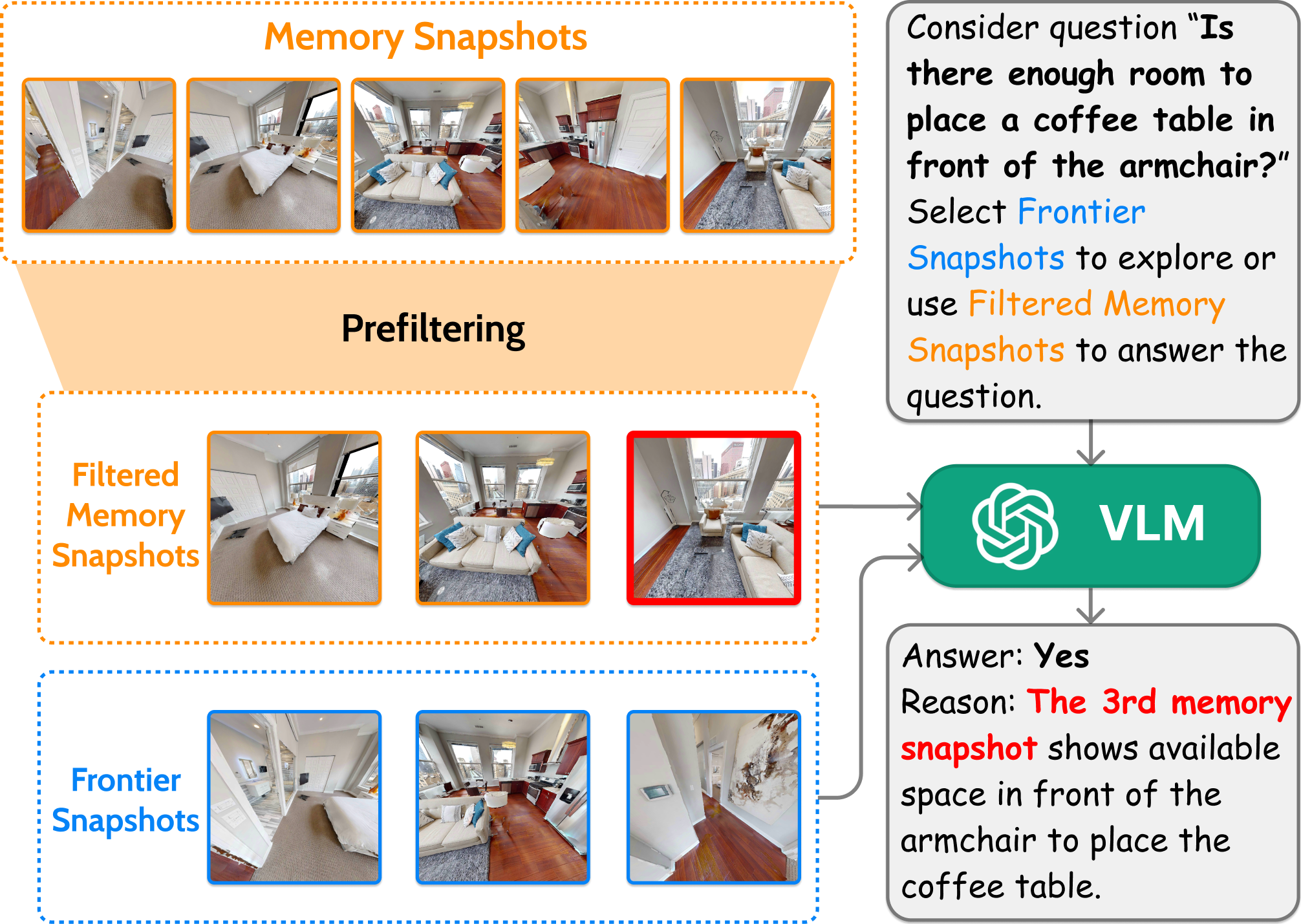}
    \caption{\textbf{3D-Mem as visual input for the VLM in embodied question answering.} The VLM first retrieves relevant memory snapshots with prefiltering, then utilizes the frontier snapshots and memory snapshots to perceive the scene and reason about the embodied questions.}
    \label{fig:query_llm}
    \vspace{-10pt}
\end{figure}

With the updated frontier and memory snapshots, we can directly leverage the perception and reasoning capabilities of large VLMs, as the image nature of frontier and memory snapshots makes them easily interpretable by VLMs. This is a core advantage of 3D-Mem over other 3D representations such as point clouds, as recent VLMs, trained on internet-scale data, often have stronger visual capabilities than specially fine-tuned models such as 3D-LLM~\cite{3dllm}.  A pseudo-code overview of the whole exploration and reasoning framework is in Figure~\ref{fig:pseudo_exploration} in Appendix.


\noindent\textbf{Versatility in Task Applications.}
3D-Mem is versatile and can be applied to various tasks. In the case of embodied question answering (illustrated in Figure~\ref{fig:query_llm}), the VLM agent decides whether to explore a frontier or answer the question based on the current memory snapshots. If the agent chooses a frontier, it provides a rationale for exploring that direction; otherwise, it directly answers the question, concluding the exploration episode. In object navigation tasks, where the agent aims to find a specific object, we augment each memory snapshot with image crops of the objects it contains. The agent then selects an object from one of the memory snapshots to navigate toward. Detailed experiments on these two tasks are presented in Section~\ref{exp_aeqa} and ~\ref{exp_goatbench}, with the complete prompt provided in Appendix~\ref{appendix_full_prompts}.

\noindent\textbf{Navigation Strategy.} 
If the VLM agent chooses a frontier $F_i$ to explore, the agent then directly navigates towards the location of the frontier $p_i$ from its current pose. The agent stops either when it has moved a maximum distance of $1.0m$ from its original pose or when it arrives within $0.5m$ of $p_i$. We assume a collision-free planner is available to guide the agent along the shortest path within the explored region between the two locations. Experimentally, we adopt the pathfinder provided by Habitat-sim to compute such paths and move the agent incrementally at each time step. This strategy is similar to that of ExploreEQA \citep{ren2024explore}, which also uses VLMs to guide exploration.
\vspace{-2pt}
\section{Experiments}
\label{sec:exp}

As a form of 3D scene memory, 3D-Mem stores rich and compact visual information, serving as an effective framework for a lifelong agent to explore and reason about a 3D scene. To comprehensively evaluate 3D-Mem, we begin with Active Embodied Question Answering (Section~\ref{exp_aeqa}), where the scene is initially unknown. This assessment tests 3D-Mem’s overall performance in scenarios that require both embodied exploration and reasoning. Next, we examine 3D-Mem’s efficiency in representing 3D scene information through Episodic Memory Embodied Question Answering (Section~\ref{exp_emeqa}). In this evaluation, the scene scan of the ground truth region is provided and no exploration is needed for question answering.
Following this, we evaluate 3D-Mem on GOAT-Bench (Section~\ref{exp_goatbench}), a multi-modal lifelong navigation benchmark, to demonstrate 3D-Mem's effectiveness as a lifelong memory system. Finally, we conduct a series of ablation studies to determine key components and hyperparameter choices in Appendix~\ref{appendix_ablation}.

Since 3D-Mem is a versatile scene memory, we adapt it to different benchmarks in slightly different ways, as detailed in Appendix~\ref{appendix_experiment_details}. We also include more discussion and qualitative analysis in Appendix~\ref{appendix_discussion}.

\subsection{Active Embodied Question Answering}
\label{exp_aeqa}

On A-EQA~\citep{OpenEQA2023} benchmark (Table~\ref{tab:a-eqa}), we evaluate 3D-Mem's ability to dynamically construct scene representations for exploration and reasoning given complex questions. More implementation details are in Appendix~\ref{appendix_experiment_details_aeqa}.

\begin{table}[ht]
    \centering
    \resizebox{\linewidth}{!}{
    \begin{tabular}{llcc}
        \toprule
        \textbf{} & \textbf{Method} & \textbf{LLM-Match $\uparrow$} & \textbf{LLM-Match SPL $\uparrow$} \\
        \midrule
        & \textbf{\textit{Blind LLMs}} \\
        & GPT-4* & 35.5 & N/A \\
        & GPT-4o$^{\dagger}$ & 35.9 & N/A \\
        \midrule
        & \textbf{\textit{Question Agnostic Exploration}} \\
        & CG Scene-Graph Captions* & 34.4 & 6.5 \\
        & SVM Scene-Graph Captions* & 34.2 & 6.4 \\
        & LLaVA-1.5 Frame Captions* & 38.1 & 7.0 \\
        & Multi-Frame*$^{\dagger}$ & 41.8 & 7.5 \\
        \midrule
        & \textbf{\textit{VLM Exploration}} \\
        & Explore-EQA$^{\dagger}$ & 46.9 & 23.4 \\
        & CG w/ Frontier Snapshots$^{\dagger}$ & 47.2 & 33.3 \\
        & 3D-Mem (Ours)$^{\dagger}$ & \textbf{52.6} & \textbf{42.0} \\
        \midrule
        & Human Agent* & 85.1 & N/A \\
        \bottomrule
    \end{tabular}
    }
    \caption{\textbf{Experiments on A-EQA.} ``CG" denotes ConceptGraphs. Methods with * are reported from OpenEQA~\citep{OpenEQA2023}, and with ${\dagger}$ are evaluated on the 184-question subset.
    }
    \label{tab:a-eqa}
\end{table}

\noindent\textbf{Benchmark.}
A-EQA consists of 557 questions drawn from 63 scenes in HM3D \citep{ramakrishnan2021habitat}. Due to resource limitations, our evaluation focuses on a subset of 184 questions, as mentioned in the OpenEQA benchmark~\citep{OpenEQA2023}. We also include the evaluation of 3D-Mem on the full set in Appendix~\ref{appendix_full_evaluation}.
The open-vocabulary and open-ended questions in A-EQA encompass diverse daily tasks such as object recognition, functional reasoning, and spatial understanding. For each question, an agent is initialized at a specific location and is required to explore the scene to gather the necessary information for answering the question. 

\noindent\textbf{Metrics.}
Following OpenEQA, we employ LLM-Match and LLM-Match SPL for quantitative evaluation. We first rate each predicted answer from 1 to 5 using GPT-4 to compare ground-truth and predicted answers. Given the predicted answers, LLM-Match, which measures the answer accuracy, is calculated as the average score for each question, mapped to a 0-100 scale. LLM-Match SPL, which measures the exploration efficiency, is then calculated by weighting the LLM-Match score by exploration path length. For the questions where the VLM Exploration methods failed to provide an answer, we ask GPT-4o to directly guess an answer without visual inputs, setting the SPL to 0.0.

\noindent\textbf{Baselines.}
For baselines that use VLM for exploration, we mainly compare 3D-Mem with Explore-EQA~\citep{ren2024explore} and ConceptGraph~\citep{conceptgraph} w/ frontier snapshots. We adapt Explore-EQA for open-ended questions by halting exploration and answering the question with the egocentric view once the VLM's confidence in the question exceeds a predetermined threshold. We integrate ConceptGraph into our exploration pipeline by replacing memory snapshots with object image crops, while maintaining other settings the same, including prefiltering and how answers are obtained. We adopt GPT-4o as the choice of VLM by directly utilizing the OpenAI API.
Besides the methods that can do active exploration above, we also include other simple baselines implemented by OpenEQA. The group of question-agnostic exploration baselines employ question-agnostic frontier exploration to obtain an episodic memory of image frames. These frames are subsequently used to prompt VLMs directly (Multi-Frame), generate frame captions as prompts for LLMs (LLaVA-1.5 Frame-Captions), or construct textual scene-graph representation using ConceptGraph (CG) and Sparse Voxel Map (SVM) to prompt LLMs. Additionally, blind LLM experiments are included, where the LLM is tasked with answering questions without any visual information. Note that the Multi-Frame baseline uses 75 frames for each question, and is evaluated on the 184-question subset. Other baselines from OpenEQA are evaluated  on the full 557-question set. 

\noindent\textbf{Analysis.}
As shown in Table~\ref{tab:a-eqa}, 3D-Mem significantly outperforms previous methods in both accuracy and efficiency. The major takeaways are as follows: 1) The superior performance in open-ended embodied question answering highlights the advantages of using snapshots as a memory format, which store richer and more flexible visual information for the VLM to address complex questions. In contrast, object-based memory systems, which use either image crops or language captions to represent objects and spatial relationships, are less robust when handling diverse questions, as they rely on rigid object-level features. 2) The multi-frame VLM implemented by OpenEQA also achieves inferior results, despite using a similar multi-view imagesrepresentation, because this baseline linearly selects episodic frames and includes excessive repetitive or irrelevant information for the questions. In contrast, 3D-Mem creates an average of 10.94 memory snapshots from 39.76 egocentric observations after each episode. After prefiltering with $K=10$, only 3.26 memory snapshots are retained as direct input for VLMs. These results demonstrate the compactness and efficiency of 3D-Mem as scene memory.

\vspace{-2mm}
\subsection{Episodic-Memory Embodied Q\&A}
\label{exp_emeqa}
\vspace{-2mm}
We evaluate the representation capability of 3D-Mem on EM-EQA~\citep{OpenEQA2023} to further demonstrate 1) the effectiveness of image memory compared to captions, 2) the compact and informative nature of our method. More implementation details are in Appendix~\ref{appendix_experiment_details_emeqa}.

\begin{table}[]
    \centering
    \resizebox{0.7\linewidth}{!}{
    \begin{tabular}{llcc}
        \toprule
        \textbf{Methods} & \textbf{Avg. Frames} & \textbf{LLM-Match} \\
        \midrule
        Blind LLM*         & 0              & 35.5\\ 
        CG Captions* & 0 & 34.4 \\
        SVM Captions* & 0 & 34.2 \\
        Frame Captions* & 0 & 38.1 \\
        Multi-Frame & 3.0 & \underline{48.1} \\
        3D-Mem (Ours) & 3.1 & \textbf{57.2} \\
        \midrule
        Human                & Full           & 86.8  \\ 
        \bottomrule
    \end{tabular}
    }
    \caption{\textbf{Experiments on EM-EQA.} Frame Efficiency and performance. Methods denoted by * use GPT-4 to generate answers, as reported in OpenEQA}
    \vspace{-0.5cm}
    \label{tab:em-eqa}
\end{table}

\begin{figure}
    \centering
    \includegraphics[width=0.8\linewidth]{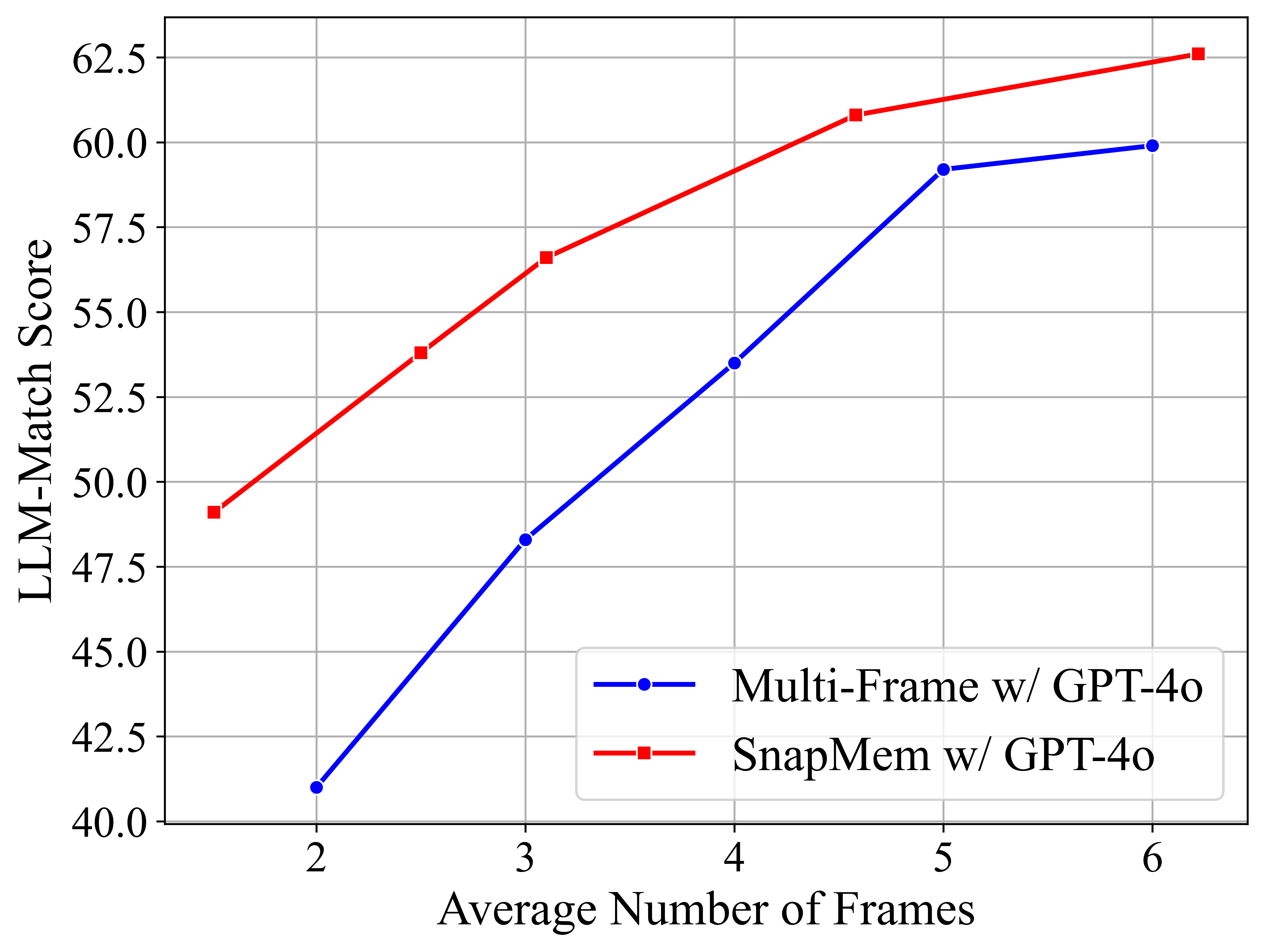}
    \caption{\textbf{Frame Efficiency of 3D-Mem on EM-EQA.} LLM-Match Score vs. Average Number of Frames for 3D-Mem and Multi-Frame both using GPT-4o}
    \label{fig:em-eqa}
    \vspace{-3mm}
\end{figure}


\noindent\textbf{Benchmark.}
EM-EQA is an Embodied Q\&A benchmark that contains over 1600 questions from 152 ScanNet~\citep{dai2017scannet} and HM3D\citep{ramakrishnan2021habitat} scenes. The open-vocabulary and open-ended questions in EM-EQA encompass diverse daily tasks such as object recognition, functional reasoning, and spatial understanding. For each question, a trajectory comprising RGB-D observations and the corresponding camera poses at each step is provided, offering necessary contextual information needed to answer the questions.

\noindent\textbf{Baselines.}
We compare against language-only scene representations, including ConceptGraphs captions, Sparse Voxel Maps Captions, and Frame Captions. We also compare against Multi-Frame, which directly processes 2 to 6 linearly sampled frames using GPT-4o.

\noindent\textbf{Analysis.}
From the results, we can observe that: 1)
As shown in Table~\ref{tab:em-eqa}, both 3D-Mem and Multi-Frame significantly outperform methods that rely on captions to represent a 3D scene while using only approximately three frames. This demonstrates the effectiveness of using a set of images to represent a 3D scene and highlights the limitations of 3D scene graph captions in addressing complex queries involving relationships between objects. 2) 
In both Table~\ref{tab:em-eqa} and Figure~\ref{fig:em-eqa}, 3D-Mem surpasses Multi-Frame in frame efficiency, underscoring the compact and informative nature of our proposed 3D scene memory.

\vspace{-2mm}
\subsection{GOAT-Bench}
\label{exp_goatbench}
\vspace{-2mm}
On GOAT-Bench~\citep{khanna2024goatbench} (Table~\ref{tab:goat-bench}), we evaluate 3D-Mem's effectiveness as a lifelong memory system that facilitates efficient exploration and reasoning. More implementation details are in Appendix~\ref{appendix_experiment_details_goatbench}.

\noindent\textbf{Benchmark.}
GOAT-Bench is a multimodal lifelong navigation benchmark, where an agent is tasked with sequentially navigating to several objects in an unknown scene, with each target described by either a category name (\textit{e.g.}, microwave), a language description (\textit{e.g.}, the microwave on the kitchen cabinet near the fridge), or an image of the target object. Due to the large size of GOAT-Bench and the resource limitations, we assess a 1/10-size subset of the ``Val Unseen'' split, consisting of one exploration episode for each of the 36 scenes, totaling 278 navigation subtasks. For reference, we also include the evaluation of 3D-Mem on the full test set in Appendix~\ref{appendix_full_evaluation}.

\noindent\textbf{Metrics.}
GOAT-Bench employs the Success Rate and Success weighted by Path Length (SPL), similar to A-EQA dataset. A navigation task is deemed success if the agent's final location is within 1 meter from the navigation goal. SPL is the success score weighted by exploration distances.

\noindent\textbf{Baselines.} 
Similar to the experiments in A-EQA, we compare 3D-Mem with Explore-EQA~\citep{ren2024explore} and ConceptGraph~\citep{conceptgraph} baselines. Due to implementation differences in Explore-EQA, we introduce an additional success criterion for this baseline: a subtask is considered successful if the target object is visible in the final observation. This extra criterion leverages ground truth grounding, thereby enhancing the baseline's capability. To demonstrate the effectiveness of 3D-Mem's lifelong memory, we include another baseline (3D-Mem w/o memory) in which we clear the constructed scene graph after each subtask. We also directly include baselines implemented in GOAT-Bench. However, these baselines are simple RNN-based models trained via reinforcement learning, which causes their performance to lag behind the baselines we implemented.

\noindent\textbf{Analysis.}
As shown in Table~\ref{tab:goat-bench}, 3D-Mem achieves the highest scores compared to previous methods in both accuracy and efficiency. Our major observations are as follows:
1) Even though GOAT-Bench is an object-based navigation benchmark, which is well-suited for ConceptGraph settings, 3D-Mem still outperforms ConceptGraph w/ frontier snapshots. This can be attributed 3D-Mem's multi-view images representation, which captures more comprehensive information, making it easier to match with the diverse descriptions in GOAT-Bench. 
2) When compared with the original 3D-Mem, the performance of 3D-Mem w/o memory declines for both GPT-4o and LLaVA-7B models, particularly in efficiency (SPL), indicating that 3D-Mem is beneficial as a memory system for lifelong learning. 3) Explore-EQA, which uses a traditional value map for each subtask to indicate regions of interest, also performs worse, as it lacks the mechanism to memorize information in explored regions. 4) In terms of efficiency, after each episode, 3D-Mem generates an average of 16.58 memory snapshots from 91.37 egocentric observations, and after prefiltering with $K=10$, only 4.66 memory snapshots are kept for each query.

\begin{table}[ht]
    \centering
    \resizebox{\linewidth}{!}{
    \begin{tabular}{llcc}
        \toprule
        \textbf{} & \textbf{Method} & \textbf{Success Rate $\uparrow$} & \textbf{SPL $\uparrow$} \\
        \midrule
        & \textbf{\textit{GOAT-Bench Baselines}} \\
        & Modular GOAT* & 24.9 & 17.2 \\
        & Modular CLIP on Wheels* & 16.1 & 10.4 \\
        & SenseAct-NN Skill Chain* & 29.5 & 11.3 \\
        & SenseAct-NN Monolithic* & 12.3 & 6.8 \\
        \midrule
        & \textbf{\textit{Open-Sourced VLM Exploration}} \\
        & 3D-Mem w/o memory$^{\dagger}$ & 40.6 & 14.6 \\ 
        & 3D-Mem (Ours)$^{\dagger}$  & 49.6 & 29.4 \\ 
        \midrule
        & \textbf{\textit{GPT-4o Exploration}} \\
        & Explore-EQA$^{\dagger}$ & 55.0 & 37.9 \\
        & CG w/ Frontier Snapshots$^{\dagger}$ & 61.5 & 45.3 \\
        & 3D-Mem w/o memory$^{\dagger}$ & 58.6 & 38.5 \\
        & 3D-Mem (Ours)$^{\dagger}$ & \textbf{69.1} & \textbf{48.9} \\
        \bottomrule
    \end{tabular}
    }
    \caption{\textbf{Experiments on  GOAT-Bench.} Evaluated on the ``Val Unseen" split. ``CG'' denotes ConceptGraphs. Methods denoted by * are from GOAT-Bench, and with $\dagger$ are evaluated on the subset.}
    \label{tab:goat-bench}
    \vspace{-10pt}
\end{table}
\vspace{-2mm}
\section{Conclusion}
\label{sec:conclusion}
\vspace{-2mm}
We present 3D-Mem, a 3D scene memory framework that uses a set of informative multi-view snapshot images to store robust visual information of a 3D scene. With the integration of the frontier-based exploration framework, 3D-Mem allows the agent to either leverage the memory of explored regions to solve tasks or further explore the scene to expand its knowledge. With its incremental construction and efficient memory retrieval mechanism, 3D-Mem serves as an effective memory system for lifelong agents. Extensive experiments demonstrate the significant advantages of 3D-Mem over traditional scene representations.


\clearpage
\setcounter{page}{1}
\maketitlesupplementary



\section{Full-Set Evaluation}
\label{appendix_full_evaluation}
Following the common practice and due to resource limitations, we only evaluate baselines and our method on a subset of A-EQA and GOAT-Bench in our main paper. For reference, we also evaluate 3D-Mem on the complete benchmarks, as shown in the following table.

\begin{table}[ht]
\renewcommand{\arraystretch}{0.5}
    \vspace{-0.9em}
    \centering
    \resizebox{\linewidth}{!}{
    \begin{tabular}{lcc|cc}
        \toprule
        & \multicolumn{2}{c}{\textbf{Whole Set}} 
        & \multicolumn{2}{c}{\textbf{Subset}} \\
        \cmidrule(lr){2-3} \cmidrule(lr){4-5}
        \textbf{GOAT-Bench} 
        & \textbf{   Success Rate $\uparrow$} 
        & \textbf{SPL $\uparrow$}
        & \textbf{Success Rate $\uparrow$} 
        & \textbf{SPL $\uparrow$} \\
        \midrule
        3D-Mem (Ours)     
        & 62.9    & 44.7    
        & 69.1 & 48.9 \\
        \bottomrule
    \end{tabular}
    }
    \vspace{-1.0em}
    \label{tab:goat-bench-subset}
\end{table}
\begin{table}[ht]
\renewcommand{\arraystretch}{0.5}
    \centering
    \vspace{-1.3em}
    \resizebox{\linewidth}{!}{
    \begin{tabular}{lcc|cc}
        \toprule
        \multicolumn{1}{c}{} & 
        \multicolumn{2}{c}{\textbf{Whole Set}} & 
        \multicolumn{2}{c}{\textbf{Subset}} \\
        \cmidrule(lr){2-3} \cmidrule(lr){4-5}
        \textbf{A-EQA} & 
        \textbf{LLM-Match $\uparrow$} & 
        \textbf{LLM-Match SPL $\uparrow$} & 
        \textbf{LLM-Match $\uparrow$} & 
        \textbf{LLM-Match SPL $\uparrow$} \\
        \midrule
        3D-Mem (ours)     & 53.3    & 38.0   & 52.6 & 42.0 \\
        \bottomrule
    \end{tabular}
    }
    \vspace{-1.3em}
    \label{tab:a-eqa-subset}
\end{table}

\section{Discussion}
\label{appendix_discussion}

\subsection{Detailed Experiment Results}



    


    



\noindent\textbf{A-EQA.}
Table \ref{tab:detail-a-eqa} presents a detailed breakdown of results on A-EQA across the seven OpenEQA question categories. As demonstrated in the table, 3D-Mem significantly outperforms ConceptGraph w/ Frontier Snapshots for questions requiring spatial reasoning, including spatial understanding and object localization where the relative positions of surroundings is needed to generate better answers. Such performance gain is attributed to Memory Snapshot, which visually stores both the foreground inter-object spatial relationships and background room-level spatial cues. In contrast, ConceptGraph relies solely on object-centric representations, limiting its ability to capture broader spatial context. For other question categories focus on identifying object-specific variables, i.e., object recognition, attribute recognition, object state recognition, or heavily rely on external knowledge embedded within VLMs, i.e., world knowledge, 3D-Mem also showcases comparable performance as it ensures the capture of all informative objects and effectively utilizes the capability of VLMs.


Compared with Explore-EQA, 3D-Mem generally exhibits higher LLM-Match scores in object-related question categories as we explicitly represents major objects within the scene by Memory Snapshots, enabling the agent to concentrate on relevant elements that may contribute to the final answer. On the other hand, Explore-EQA has consistently lower SPL due to its inefficient semantic-map-based exploration mechanism where explicit visual information of frontiers is encoded into an implicit semantic map. 3D-Mem addresses this limitation by visually capturing glimpses of unexplored areas with Frontier Snapshots and integrating them with Memory Snapshots in the decision-making phase, which provides a more intuitive and holistic view, enabling it to make more informed and effective choices during exploration. 

\noindent\textbf{GOAT-Bench.}
Table \ref{tab:detail-goat-bench} presents a detailed breakdown of results on GOAT-Bench across the three question modalities. Comparing 3D-Mem with CG w/ Frontier Snapshots, we observe that 3D-Mem significantly outperforms CG in the Object Category and Language modalities. This improvement is attributed to 3D-Mem's memory snapshots, which provide explicit spatial relationships among objects and their surroundings, enabling the agent to locate targets more effectively. The detailed spatial context captured in the snapshots enhances the agent's ability to interpret instructions that rely on spatial cues. In contrast, 3D-Mem's SPL in the Image modality is slightly lower than CG's, despite a similar Success Rate. This decrease is likely due to current vision-language models (VLMs) struggling to relate images of complex scenes taken from different angles. When the memory snapshots and the image prompts depict the same region from different perspectives, the VLM may become distracted, leading to less efficient navigation paths. This may highlight a limitation in current VLMs' ability to match images across varying viewpoints in complex environments.

3D-Mem consistently outperforms both methods across all modalities in terms of Success Rate and SPL scores. By enabling the agent to recall previously observed regions and objects, memory significantly enhances the effectiveness and efficiency of exploration and reasoning. These results highlight memory's essential role in lifelong object navigation tasks.

\begin{table*}[ht]
    \centering
    \resizebox{\linewidth}{!}{
    \begin{tabular}{llcccccccccccccccc}
        \toprule
        \textbf{} & \textbf{Method} & \multicolumn{2}{c}{\makecell{object\\recognition}} & \multicolumn{2}{c}{\makecell{object\\localization}} & \multicolumn{2}{c}{\makecell{attribute\\recognition}} & \multicolumn{2}{c}{\makecell{spatial\\understanding}} & \multicolumn{2}{c}{\makecell{object state\\recognition}} & \multicolumn{2}{c}{\makecell{functional\\reasoning}} & \multicolumn{2}{c}{\makecell{world\\knowledge}} & \multicolumn{2}{c}{overall} \\
        \midrule
        & \textbf{\textit{Blind LLMs}} \\
        & GPT-4* & 25.3 & - & 28.4 & - & 27.3 & - & 37.7 &- & 47.2 &- & 54.2 & - & 29.5 & - & 35.5 & - \\
        & GPT-4o & 22.0 & - & 25.0 & - & 27.3 & - & 40.8 &- & 50.9 &- & 61.8 & - & 38.4 & - & 35.9 & - \\
        \midrule
        & \textbf{\textit{Question Agnostic Exploration}} \\
        & CG Scene-Graph Captions* & 25.3 & - & 16.5 & - & 29.2 & - & 37.0 & - & 52.2 & - & 46.8 & - & 37.8 & - & 34.4 & 6.5 \\
        & SVM Scene-Graph Captions* & 29.0 & - & 17.2 & - & 31.5 & - & 31.5 & -  & 54.2 & - & 39.8 & - & 38.9 & - & 34.2 & 6.4 \\
        & LLaVA-1.5 Frame Captions* & 25.0 & - & 24.0 & - & 34.1 & - & 34.4 & -  & 56.9 & - & 53.5 & - & 40.6 & - & 38.1 & 7.0 \\
        & Multi-Frame* & 34.0 & - & 34.3 & - & 51.5 & - & 39.5  & - & 51.9 & - & 45.6  & - & 36.6 & - & 41.8 & 7.5 \\
        \midrule
        & \textbf{\textit{VLM Exploration}} \\
        & Explore-EQA & 44.0 & 19.6 & 37.1 & 29.6 & \textbf{55.3} & 36.0 & 42.1 & 6.6 & 46.3 & 9.2 & 63.2 & 35.7 & 45.5 & 22.0 & 46.9 & 23.4 \\
        & CG w/ Frontier Snapshots & 45.0 & 42.0 & 32.1 & 25.0 & 50.8 & 35.2 & 32.9 & 18.7 & 68.5 & 38.4 & 58.8 & 42.2 & 45.5 & 33.5 & 47.2 & 33.3\\
        & 3D-Mem (Ours) & \textbf{49.0} & \textbf{45.2} & \textbf{48.6} & \textbf{41.3} & 47.7 & \textbf{38.6} & \textbf{43.4} & \textbf{33.3}  & \textbf{69.4} & \textbf{50.3} & \textbf{64.7} & \textbf{47.2} & \textbf{49.1} & \textbf{38.9} & \textbf{52.6} & \textbf{42.0} \\
        \midrule
        & Human Agent* & 89.7 & - & 72.8 & - & 85.4 & - & 84.8 & - & 97.8 & - & 78.9 & - & 88.5 & - & 85.1 & - \\
        \bottomrule
    \end{tabular}
    }
    \caption{\textbf{Performance on A-EQA by Question Categories.} For each question categories, there are two columns. The first column stands for the LLM-Match Score, while the second column represents the SPL score. ``CG" denotes ConceptGraphs. Methods with * are reported from OpenEQA~\citep{OpenEQA2023}. Columns represent different category of questions in the dataset.
    }
    \label{tab:detail-a-eqa}
\end{table*}

\begin{table*}[ht]
    \centering
    \resizebox{\linewidth}{!}{
    \begin{tabular}{llcccccccc}
        \toprule
        \textbf{} & & \multicolumn{2}{c}{Object Category} & \multicolumn{2}{c}{\makecell{Language}} & \multicolumn{2}{c}{Image} & \multicolumn{2}{c}{Overall}\\
        \cmidrule(r){3-4} \cmidrule(r){5-6} \cmidrule(r){7-8} \cmidrule(r){9-10} 
        \textbf{} & \textbf{Method} & \textbf{Success Rate} & \textbf{SPL} & \textbf{Success Rate} & \textbf{SPL} & \textbf{Success Rate} & \textbf{SPL} & \textbf{Success Rate} & \textbf{SPL} \\
        \midrule
        & \textbf{\textit{Open-Sourced VLM Exploration}} \\
        & 3D-Mem w/o memory & 55.6 & 16.0 & 33.3 & 15.5 & 31.8 & 12.2 & 40.6 & 14.6 \\ 
        & 3D-Mem (Ours)  & 62.6 & 33.3 & 49.5 & 31.7 & 35.2 & 22.7 & 49.6 & 29.4\\ 
        \midrule
        & \textbf{\textit{GPT-4o Exploration}} \\
        & Explore-EQA & 64.7 & 48.4 & 42.9 & 22.7 & 56.8 & 41.8 & 55.0 & 37.9\\
        & CG w/ Frontier Snapshots & 65.3 & 44.7 & 55.0 & 38.9 & 64.0 & \textbf{52.8} & 61.5 & 45.3\\
        & 3D-Mem w/o memory & 69.9 & 45.4 & 50.35 & 30.1 & 54.4 & 39.5 & 58.6 & 38.5\\
        & 3D-Mem (Ours) & \textbf{79.2} & \textbf{55.8} & \textbf{61.9} & \textbf{46.0} & \textbf{65.2} & 44.2 & \textbf{69.1} & \textbf{48.9}\\
        \bottomrule
    \end{tabular}
    }
    \caption{\textbf{Performance on  GOAT-Bench by Question Modalities.} Evaluated on the ``Val Unseen" split. ``CG'' denotes ConceptGraphs. Methods denoted by * are from GOAT-Bench.}
    \label{tab:detail-goat-bench}
    \vspace{-10pt}
\end{table*}

\subsection{Decision Frequency}
Experimentally, the agent queries the VLM and makes a new decision after moving 
$1m$ towards the target. We also tested an alternative approach where the agent makes a new decision only after reaching the navigation target. For example, if the agent selects a frontier, it navigates directly to that frontier's location before making its next choice. However, this approach results in LLM-match scores and SPL values of 50.5 and 36.2, respectively, on A-EQA, which are suboptimal particularly for SPL. Under this setting, we observed numerous cases where the agent initially selects an incorrect frontier and must fully navigate to it before revising its decision, leading to significant wasted exploration distance. In contrast, with our default setting, the agent can adjust its decision en route, mitigating such inefficiencies. One advantage of the alternative setting, however, is that it prevents the agent from oscillating between two frontiers, a problem that can arise in our default setting, particularly during longer exploration episodes. For this reason, in our demo, we opted to have the agent navigate to the target before making the next decision.

\subsection{Limitations}
We acknowledge several key constraints of 3D-Mem: 
(1) Similar to most 3D scene representations, 3D-Mem is designed for static environments and is not robust to moving objects.
(2) The performance of 3D-Mem depends on object detection results and VLM reasoning capabilities.
(3) The precise location of the agent is required to accurately locate the objects in the scene and construct scene memory, which could be challenging to acquire after long-time exploration.
(4) The latency of the whole pipeline during embodied QA is still noticeable. We time the detailed latency of each component of our pipeline in Table~\ref{tab:latency}. We also argue that, although 2D-3D lifting (including SLAM and object detection) and Prefiltering consume considerable time, in real-world scenarios, they run concurrently during navigation, and caching Prefiltering results minimizes their impact on throughput. The primary bottleneck is VLM inference, which is caused by both heavy model computation and network latency. This can be mitigated by running the model locally or optimizing VLMs with techniques such as model quantization. Additionally, 3D-Mem requires VLM inference only after each high-level step, thereby reducing the VLM query frequency.

\begin{table}[ht]
\renewcommand{\arraystretch}{0.6}
    \centering
    \resizebox{\linewidth}{!}{
    \begin{tabular}{lllccc}
        \toprule
        \textbf{} & \textbf{Component} & \textbf{2D-3D Lift} & \textbf{Clustering} & \textbf{Prefiltering} & \textbf{VLM Inference} \\
        \midrule
        & \textbf{A-EQA} & 2.43 & 0.04 & 1.12 & 3.34 \\
        & \textbf{GOAT-Bench} & 2.79 & 0.09 & 1.35 & 3.58 \\
        \bottomrule
    \end{tabular}}
    \label{tab:latency}
    \caption{Time cost of each component on A-EQA, evaluated in seconds.}
\end{table}

\section{Failure Case Analysis}

\begin{figure*}[h]
    \centering
    \includegraphics[width=0.8\linewidth]{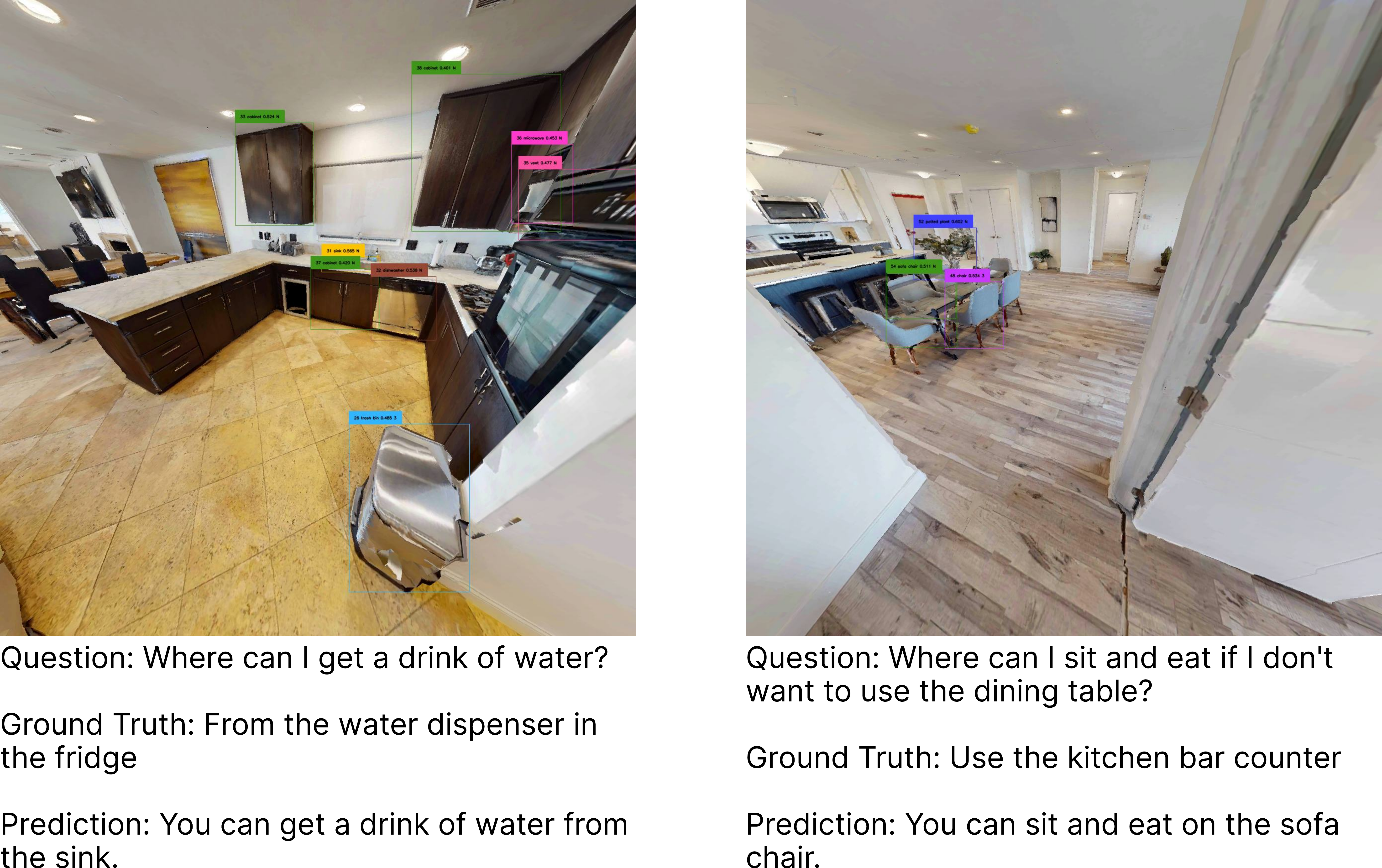}
    \caption{Failure Case 1: Some questions in A-EQA are vague and may have multiple reasonable answers.}
    \label{fig:supp_failure_1}
\end{figure*}

\begin{figure*}[h]
    \centering
    \includegraphics[width=0.8\linewidth]{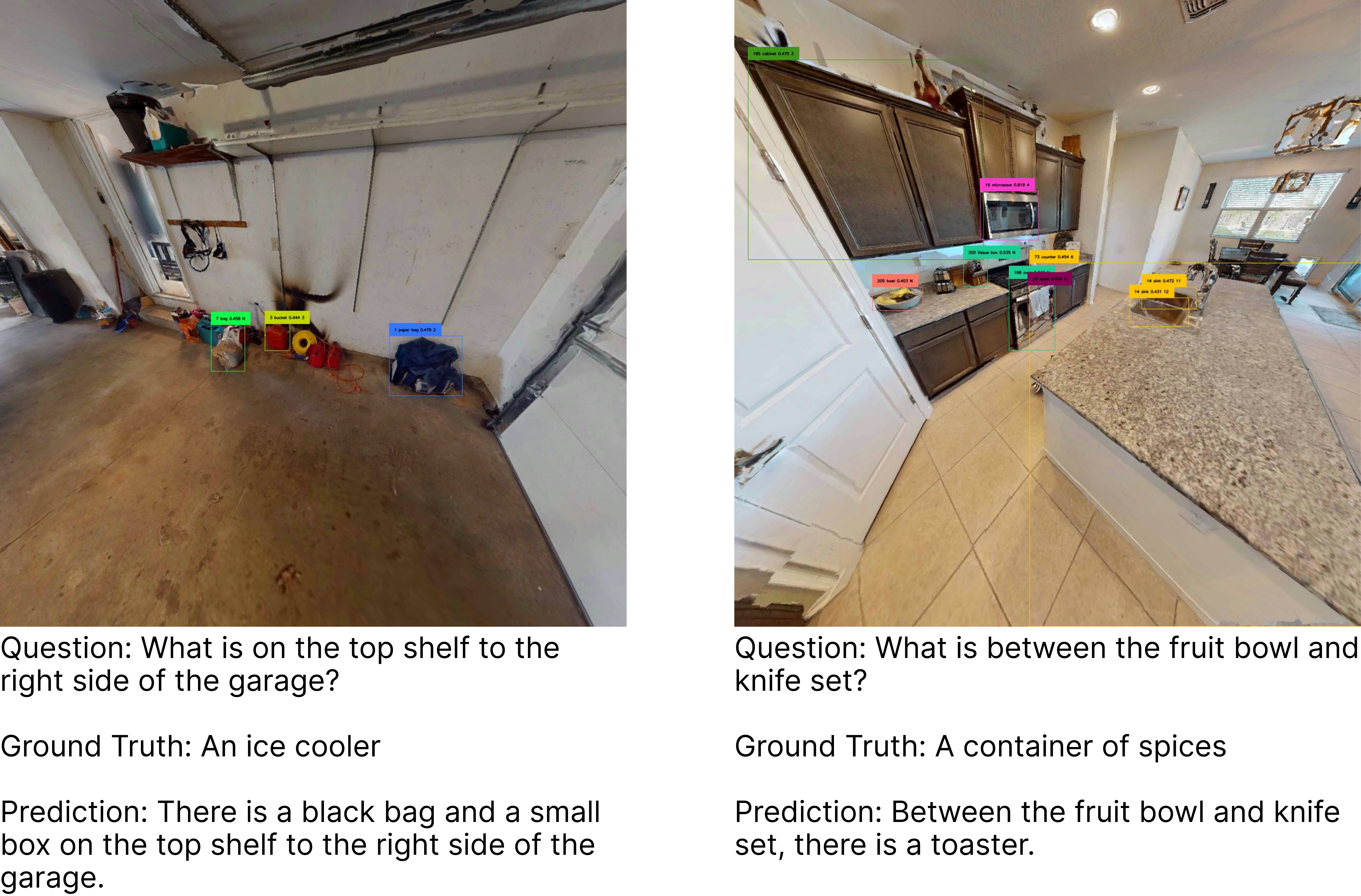}
    \caption{Failure Case 2: Due to limitations in perception capabilities and image resolution, the VLM cannot provide the correct answer even when the memory snapshot is correctly chosen.}
    \label{fig:supp_failure_2}
\end{figure*}

\begin{figure*}[h]
    \centering
    \includegraphics[width=0.8\linewidth]{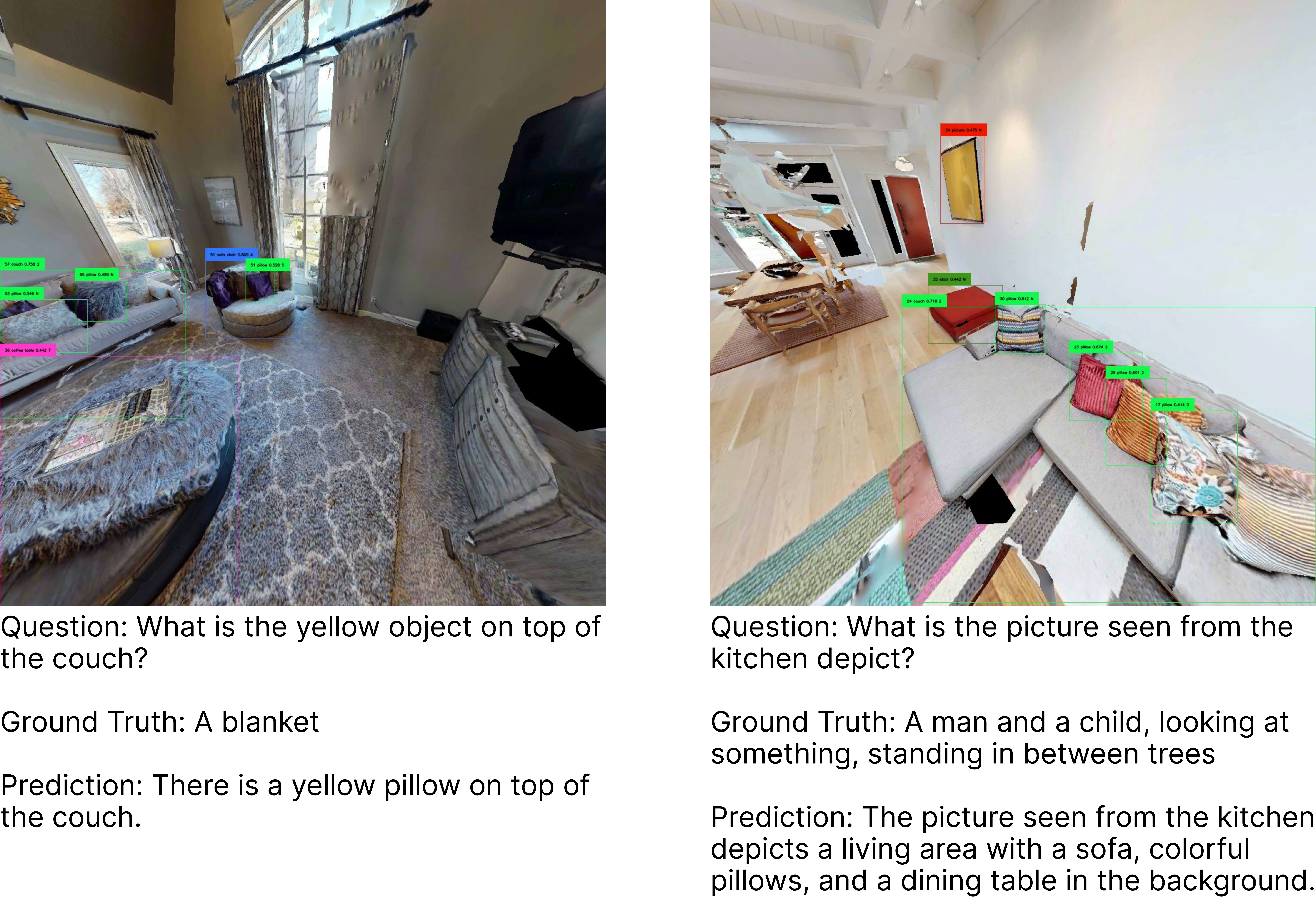}
    \caption{Failure Case 3: In some cases, the VLM selects an entirely incorrect memory snapshot and produces unreasonable answers.}
    \label{fig:supp_failure_3}
\end{figure*}

\begin{figure*}[h]
    \centering
    \includegraphics[width=0.8\linewidth]{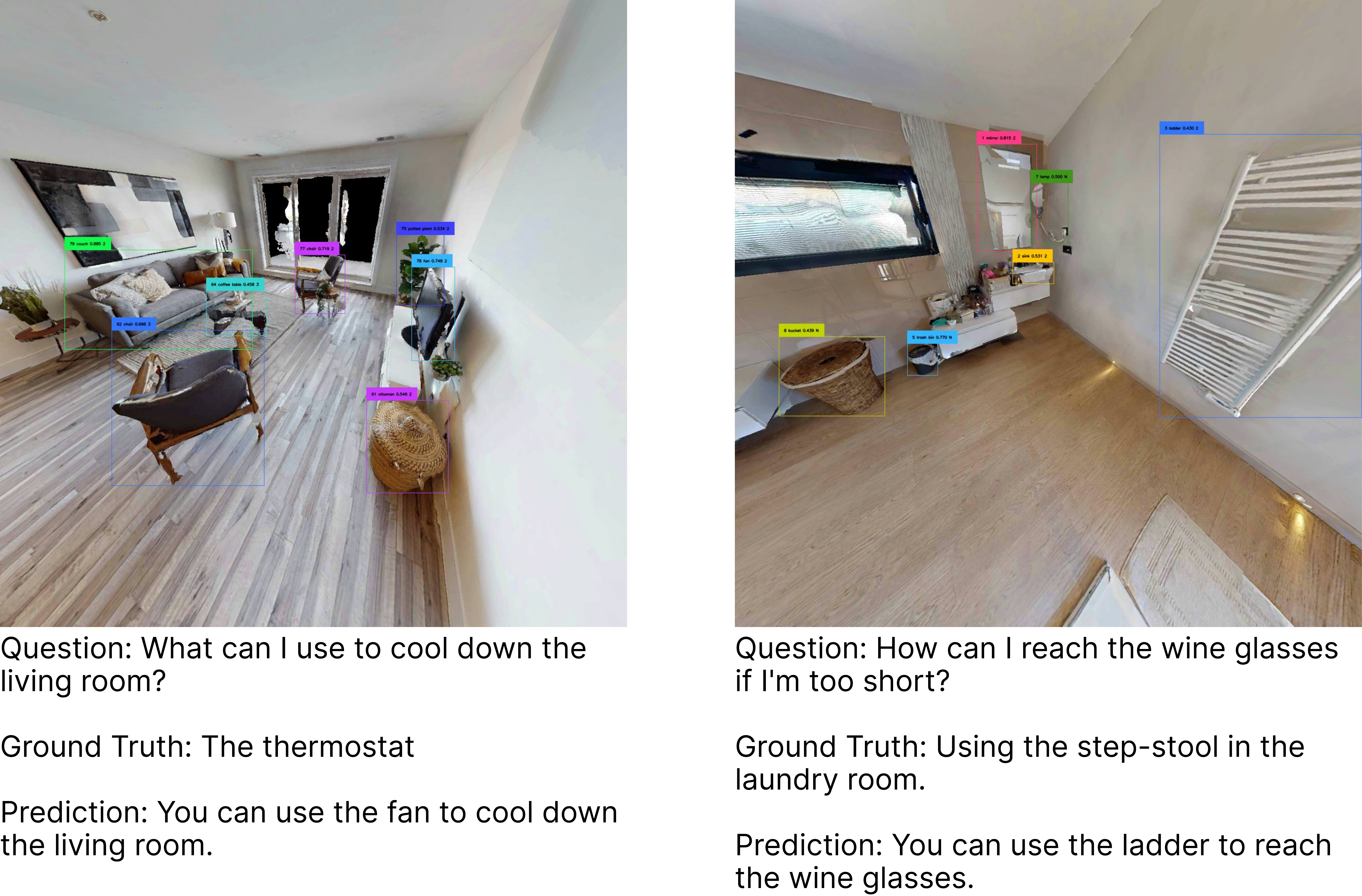}
    \caption{Failure Case 4: Incorrect labels predicted by the object detector can mislead the VLM. Although the VLM can often ignore such errors, in certain cases, these misclassifications cause confusion.}
    \label{fig:supp_failure_4}
\end{figure*}

\begin{figure*}[h]
    \centering
    \includegraphics[width=0.8\linewidth]{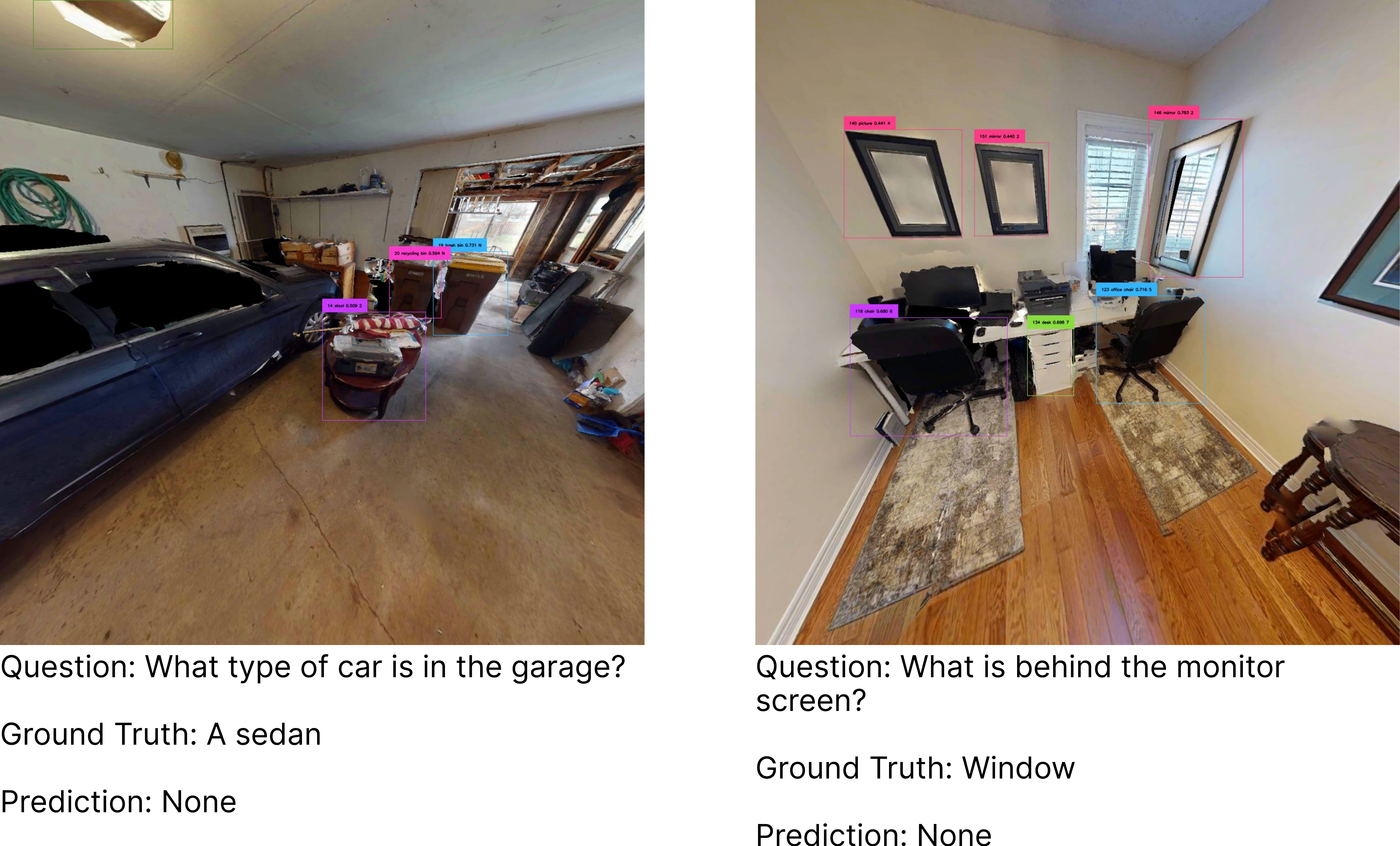}
    \caption{Failure Case 5: Some target objects are not added to the scene graph due to missed detections by the object detector. The memory snapshot shown above is where the target object should have been detected and assigned to.}
    \label{fig:supp_failure_5}
\end{figure*}

\begin{figure*}[h]
    \centering
    \includegraphics[width=0.8\linewidth]{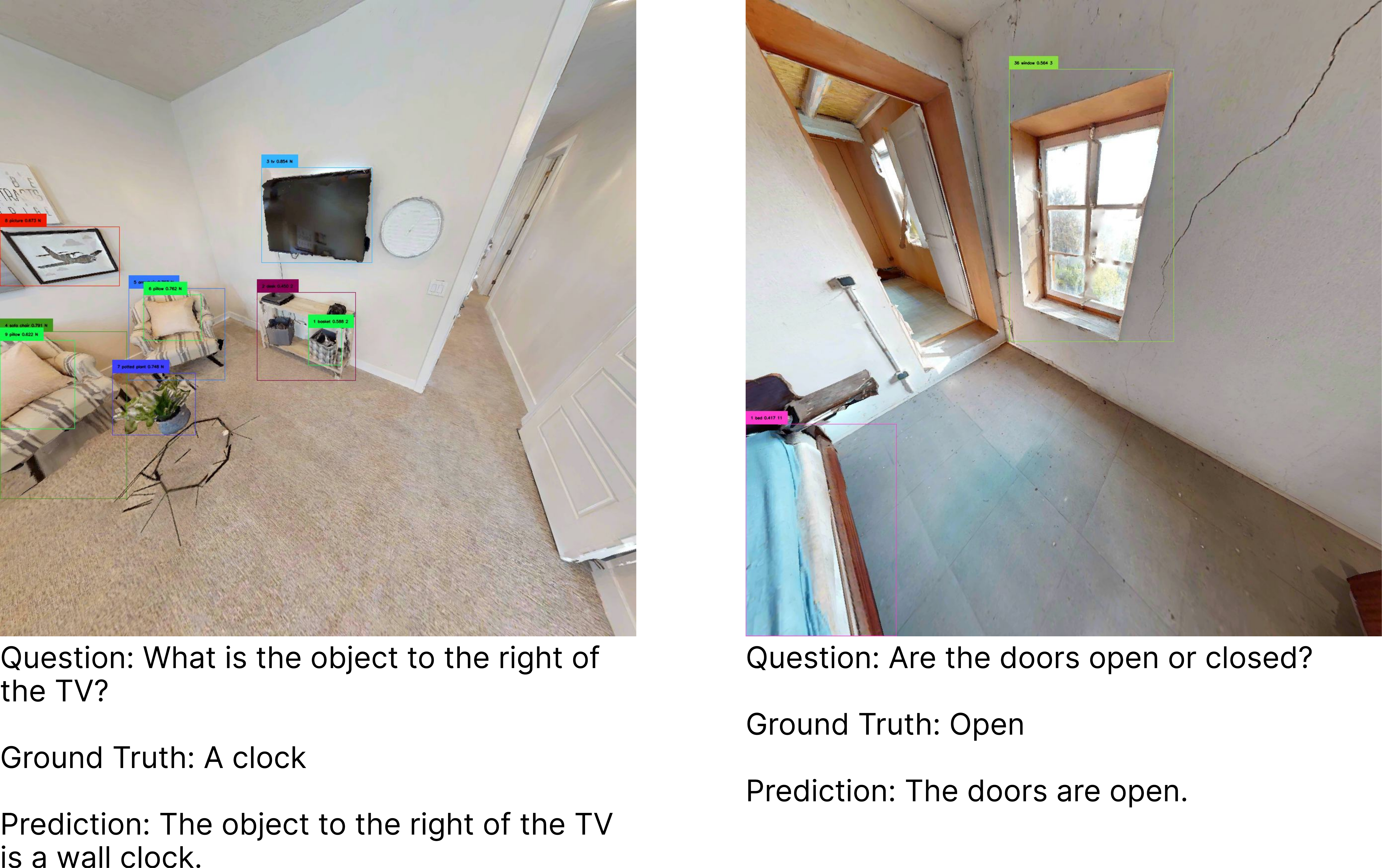}
    \caption{A similar scenario to Figure~\ref{fig:supp_failure_5}, where the target objects are not detected. However, as they are still visible in other memory snapshots, the VLM still successfully answers the questions.}
    \label{fig:supp_failure_5_additional}
\end{figure*}

In Figure~\ref{fig:supp_failure_1} to~\ref{fig:supp_failure_5_additional}, we analyze and categorize the types of questions where 3D-Mem performs poorly in A-EQA. For each example question, we provide the ground truth answer, 3D-Mem’s predicted answer, and the memory snapshot selected for answering the question. Each memory snapshot includes object detection annotations to better visualize which objects are detected and incorporated into the scene graph. These annotations can be zoomed in for a clearer view. Note that these annotations are not part of the actual input provided to the VLM.

We generally classify the failures into the following three categories:
\begin{itemize}
    \item \textbf{Dataset Issues.} As shown in Figure~\ref{fig:supp_failure_1}, some questions in the A-EQA dataset are inherently vague and allow for multiple reasonable answers. Although the ground truth answers are generally more appropriate, the VLM often exhibits overconfidence in its current predictions and terminates the episode prematurely.

    \item \textbf{Limitations of the VLM}. Many questions fail due to the limited perception capabilities of the VLM, which can be further divided into two subcategories. In the first case, as shown in Figure~\ref{fig:supp_failure_2}, the correct memory snapshot is selected, but the predicted answer is incorrect. This often occurs when the target objects in the snapshots are small, and the limited image resolution ($360 \times 360$) makes it difficult for the VLM to identify them.
    In the second case, as shown in Figure~\ref{fig:supp_failure_3}, the VLM selects the wrong memory snapshot entirely and produces unreasonable answers.

    \item \textbf{Limitations of the Object Detection Model.} 3D-Mem relies on an object detector to identify and add new objects to the scene graph. However, the object detector can sometimes produce incorrect labels, as shown in Figure~\ref{fig:supp_failure_4}. In most cases, the prefiltering process successfully filters out these incorrect labels, as they are often highly irrelevant. Additionally, the VLM is generally capable of recognizing and ignoring such errors. However, certain situations, as in Figure~\ref{fig:supp_failure_5}, illustrate cases where the detector mislabels objects—such as detecting a TV as a fan or a cloth rack as a ladder. These misclassifications can confuse the VLM, especially when the incorrect labels closely align with the expected answer.
    In addition to incorrect detections, the target objects for answering the questions are detected at all. As shown in Figure~\ref{fig:supp_failure_5}, if the car or the monitor had been detected in the relevant snapshots, the question could have been answered correctly. However, since these objects were not detected and included in memory, the VLM could not select an appropriate memory snapshot, and the agent eventually exceeded the step limit.
    Interestingly, in some cases, as shown in Figure~\ref{fig:supp_failure_5_additional}, even when the target objects are not detected, they remain visible in other memory snapshots that pass the prefiltering process. In these cases, the VLM is still able to answer the question successfully. This demonstrates that 3D-Mem is more robust than traditional 3D scene graph approaches.
    
\end{itemize}

\section{Details of Frontier-based Exploration Framework}
\label{appendix_frontier_details}
\begin{figure}[h]
    \centering
    \includegraphics[width=0.9\linewidth]{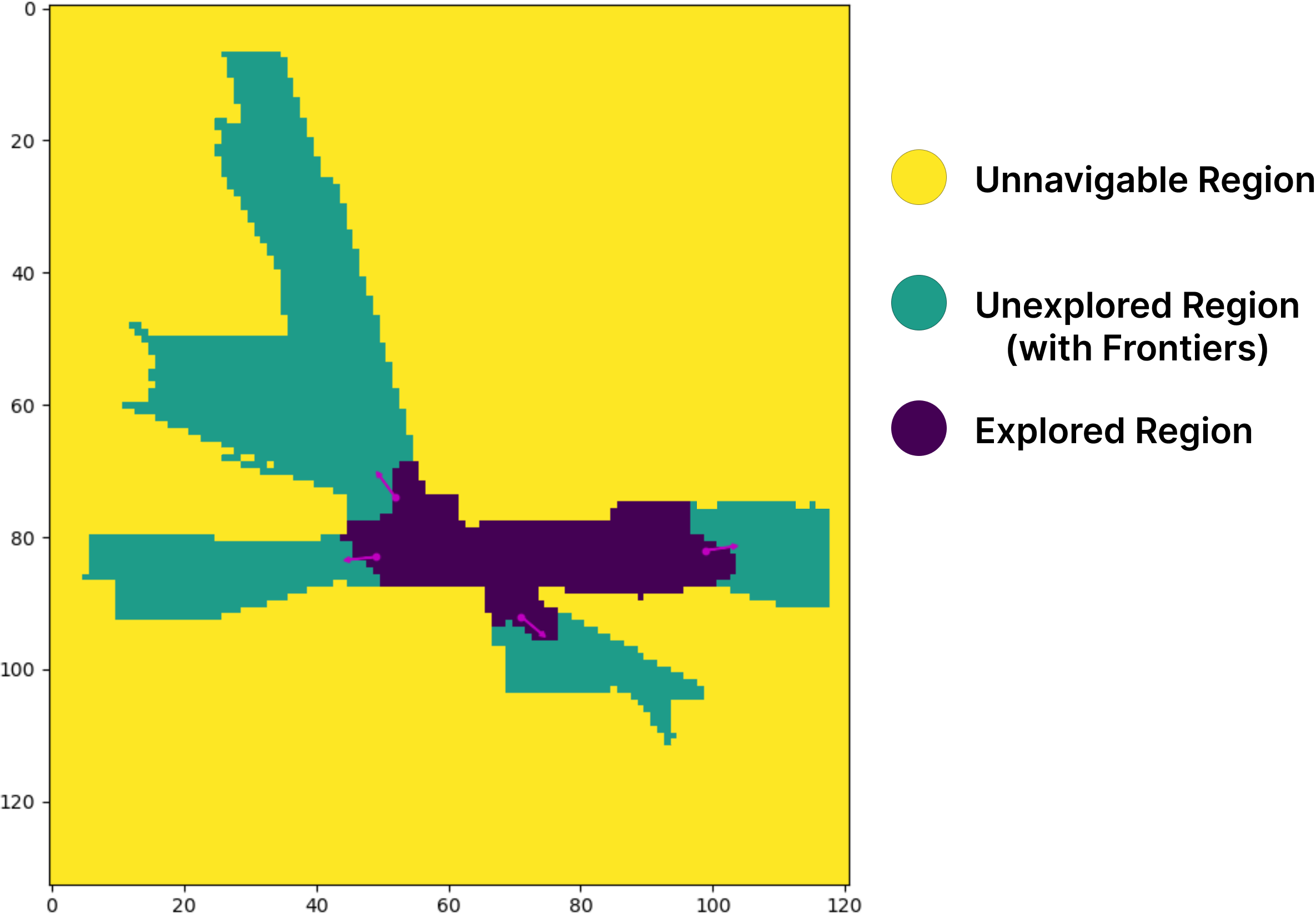}
    \caption{A illustration of different regions and frontiers in the frontier-based exploration framework. Note that navigable region consists of explored and unexplored regions.}
    \label{fig:appendix_frontier_demo}
\end{figure}

Our frontier-based exploration framework is based on the framework in Explore-EQA~\citep{ren2024explore}. We enhance its robustness and adapt it to our multi-view images representation framework. A 3D grid-based occupancy map $M$, representing the length, width and height of the entire room, is used to record the occupancy, with each voxel having a side length of 0.1 meters. During exploration, each depth observation, together with its corresponding observation pose, is used to map unoccupied spaces onto the initially fully occupied $M$. The navigable region is then defined as the layer of unoccupied voxels at the height of 0.4 meters above the ground where the agent moves. Within this navigable region, the area within 1.7 meters of the agent's trajectory is defined as the explored region, while the remainder is designated as the unexplored region, as illustrated in Figure~\ref{fig:appendix_frontier_demo}.

Frontiers are defined as clusters of pixels in the unexplored region. Pixels in the unexplored region are clustered into different groups using Density-Based Spatial Clustering of Applications with Noise (DBSCAN), with each group consisting of connected pixels. Each frontier $F = \langle r, p, I^{obs} \rangle $ represents such a pixel group $r$. The navigable location of the frontier $p$ is determined at the boundary between the frontier region and the explored region, and an image observation $I^{obs}$ is captured once the frontier has been updated. As shown in Figure~\ref{fig:appendix_frontier_demo}, each purple arrow together with a green region it points to is a frontier. For a frontier to be meaningful, $r$ must contain more than 20 pixels; otherwise, the frontier will not be created.
A frontier is considered updated if the intersection-over-union (IoU) between the new and previous regions $r$ is less than 0.95.
Additionally, if $r$ spans more than $150^{\circ}$ in the agent's field of view, it is split into two regions using K-Means clustering, resulting in two separate frontiers. This approach allows for more flexibility in choosing navigation directions. 
Also, it is important to note that this format for representing 3D space does not currently support scenes with multiple floors. Consequently, our results in Table~\ref{tab:a-eqa} fall significantly short of human performance, as many of the questions in A-EQA require exploration across different floors.

When prompting the VLM, only the image observations are included in the prompt. If the VLM chooses a frontier $F$, the location $p$ is used as the agent's navigation target.

\section{More Details in Experiments}
\label{appendix_experiment_details}

At each step $t$, we take $N=3$ egocentric views, each with a gap of $60^{\circ}$. The egocentric views are captured at a resolution of $1280 \times 1280$ for better object detection and are then resized to $360 \times 360$ as frame candidates for VLM input. Frontier snapshots are initially captured at $360 \times 360$. We use YOLOv8x-World, implemented by Ultralytics, as our detection model and a 200-class set from ScanNet~\citep{dai2017scannet} as the detection class set. We set a maximum of 50 steps for each task.

\subsection{Implementation Details for A-EQA}
\label{appendix_experiment_details_aeqa}

As explained in detail in Section~\ref{approach_integrate_with_explore}, we integrate 3D-Mem into the frontier-based exploration framework. The VLM directly returns an answer after identifying visual clues from certain memory snapshots. We set the number of egocentric observations at each step $N = 3$, the maximum distance for objects to be included in the scene graph $max\_dist = 3.5$, and the number of prefiltered classes $K = 10$.

\subsection{Implementation Details for EM-EQA}
\label{appendix_experiment_details_emeqa}

To adapt 3D-Mem to the EM-EQA benchmark, we first construct 3D-Mem for each scene using the given RGB-D observations and corresponding camera poses. For each question, we then apply prefiltering to the memory snapshots using different K values (1, 2, 3, 5, 10), and utilize the resulting filtered snapshots as prompts for GPT-4o to generate the answers.

\subsection{Implementation Details for GOAT-Bench}
\label{appendix_experiment_details_goatbench}

We reformulate the navigation task into the embodied question answering format by filling in templates for three types of target descriptions: ``Can you find the $\{$category$\}$?'', ``Can you find the object described as $\{$language description$\}$?'', and ``Can you find the object captured in the following image? $\{$image$\}$''. We adapt the prompt for navigation tasks as described in Section~\ref{approach_explore_reasoning}, allowing the VLM to choose an object directly from a memory snapshot. After the VLM identifies an object in such a way, the agent navigates to a location near that object to complete the task. We evaluate both GPT-4o and open-sourced VLM (specifically LLaVA-7B~\citep{liu2023llava}) as the choice of VLM. For LLaVA-7B model, we further fine-tune it on our generated dataset for better performance (see Appendix~\ref{appendix_finetune_llava} for more details). Other hyperparameter settings are the same as the experiments on A-EQA.

\section{Details of the Active Exploration}
\label{appendix_vlm_navigation}

\begin{figure}
    \centering
    \includegraphics[width=0.9\linewidth]{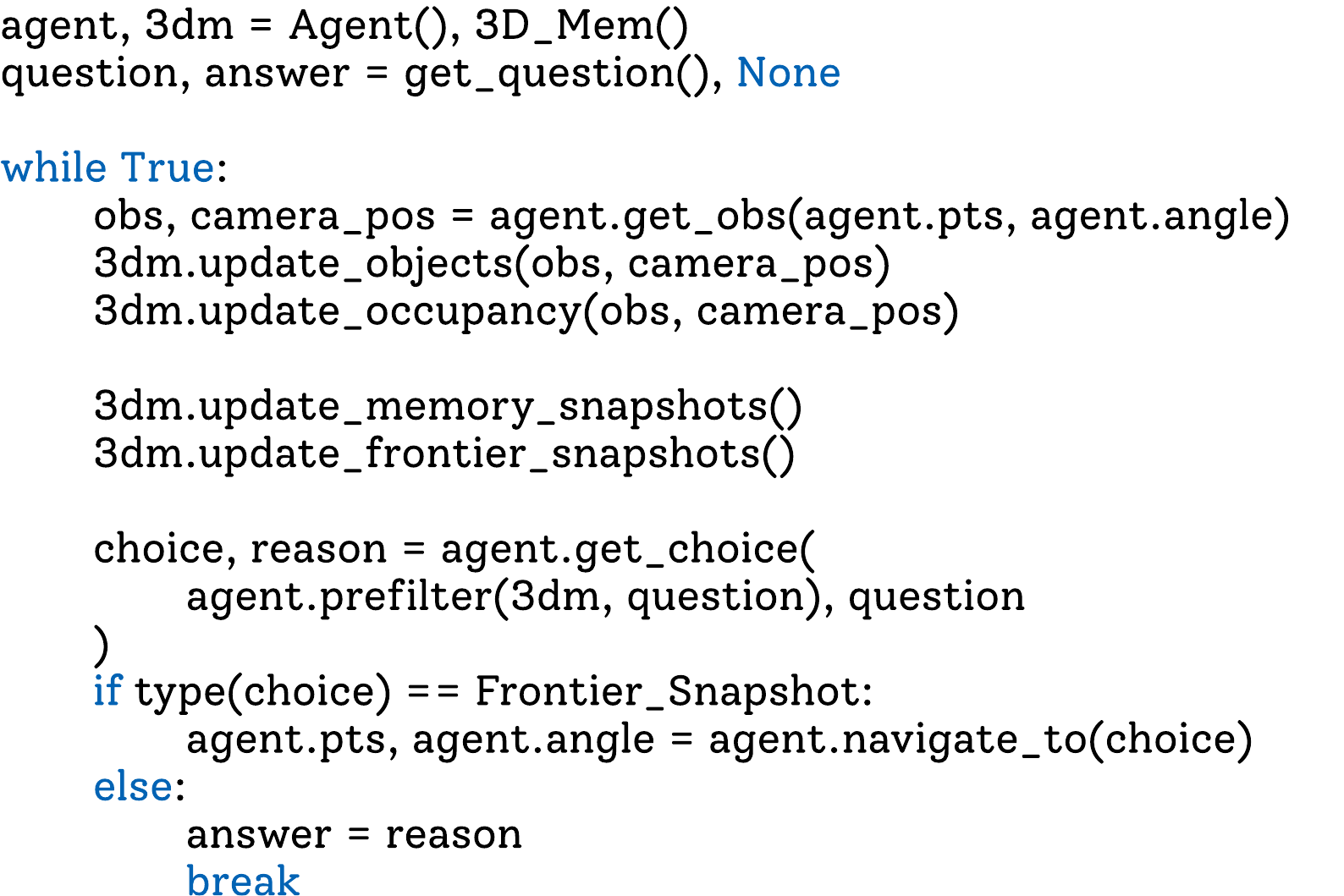}
    \caption{Overview of the frontier-based exploration pipeline with 3D-Mem on embodied question-answering task.}
    \label{fig:pseudo_exploration}
    \vspace{-3mm}
\end{figure}

When prompting the VLM for embodied question answering (A-EQA Benchmark), as shown in Figure~\ref{fig:appendix_prompt_aeqa}, we append each memory snapshot with the object classes it contains. However, we only append classes that are within the prefiltered class list. For frontier snapshots, only the raw snapshot images are used as input. The VLM will then respond with either a frontier snapshot or a memory snapshot. If the VLM returns a frontier, we set the location $p$ as the navigation target. If the VLM returns a memory snapshot along with the answer, although we directly conclude the navigation episode in our A-EQA experiments, we also set a navigation target for that memory snapshot. This allows the agent to move closer to the snapshot region, refine the selected memory snapshot, and potentially reconsider its choice.

The navigation location for a memory snapshot is determined by several conditions. We set the observation distance, $obs\_dist$, to 0.75 meters. If the snapshot contains only one object, the location is set $obs\_dist$ away from the object, in the direction from the object's location toward the center of the navigable area that is $obs\_dist$ around the object. If the memory snapshot contains two objects, the location is set $obs\_dist$ away from the midpoint of the two objects, in the direction of the perpendicular bisector of the line segment connecting the objects. If the memory snapshot contains more than two objects, we first perform Principal Component Analysis (PCA) on the object cluster to obtain the principal axis with the smallest eigenvalue. The navigation location is then set $obs\_dist$ away from the center of the object cluster, in the direction of this principal axis. Note that, in all cases for determining the navigation location, we always ignore the height of the objects and treat them as 2D points. Additionally, the above algorithm can be randomized by assigning the highest probabilities to the aforementioned positions.

Embodied navigation tasks (GOAT-Bench Benchmark) work similarly, with the following differences: 1) we append the object crop after each class name when prompting the VLM, as shown in the prompt in Figure~\ref{fig:appendix_prompt_goatbench}; 2) when the VLM returns an object choice, we treat that object as a memory snapshot containing one object and follow a similar method to set the navigation location.

After a navigation target is set (either a frontier or a memory snapshot), the agent moves towards it along a path generated by the pathfinder in habitat-sim~\citep{habitat19iccv, szot2021habitat, puig2023habitat3}. Although we utilize the pathfinder, which uses prior information from a global navmesh to find the shortest paths, we can easily replace it with a simple path-finding algorithm based on the navigable map described in Appendix~\ref{appendix_frontier_details}. Step $t$ ends after the movement. Then in the new step $t+1$, the agent updates the frontiers and memory snapshots and makes the next decision.

\section{Details of Training Open-Sourced VLMs for GOAT-Bench Navigation}
\label{appendix_finetune_llava}


\subsection{Training Dataset Collection}

In GOAT-Bench~\citep{khanna2024goatbench}, each navigation target is described by three types of descriptors: category, language, and image. We generate training data based on their provided exploration data, sourced from 136 scenes in HM3D~\citep{ramakrishnan2021habitat} training set. In each scene, a set of navigation targets is provided, each consisting of an object ID, location, category, language description, and multiple viewpoints and angles for capturing image observations. In total, the training set includes 3669 such objects, which we use as navigation targets to generate training data in our framework's format.

We adapt our exploration pipeline for data generation. For each navigation target, we first randomly select an initial point on the same floor. We then use the pathfinder in habitat-sim~\citep{habitat19iccv, szot2021habitat, puig2023habitat3} to find the shortest trajectory to the target. At each step, if the target object is present in a memory snapshot, we use that memory snapshot as the ground truth and move one step toward a location near it; if the target object is not present in any memory snapshot, we select the frontier closest to the shortest trajectory as the ground truth for that step and move one step toward that frontier. On average, we collect 4 exploration paths per target object from different initial points, with each path consisting of approximately 12 steps.

We also collect the ground truth for prefiltering by prompting GPT-4o. For each navigation target, we collect all objects that can be seen along the exploration path and feed them, together with the description, into GPT-4o. We ask GPT-4o to rank all visible objects based on their helpfulness in finding the navigation target. For each navigation target, we collect three such rankings corresponding to three types of descriptions.

\subsection{Training Process}

We fine-tune our model based on the LLaVA-1.5-7B checkpoint\citep{liu2023llava} using the collected training dataset for 5 epochs with a learning rate of 4e-6 and a batch size of 1. We use the AdamW optimizer with no weight decay. During training, DeepSpeed ZeRO-2 and LORA \citep{hu2021lora} are used to save GPU memory and accelerate training. FP16 is enabled to balance speed and precision. We train our model with 6$\times$24 Tesla V100 GPUs, and the fine-tuning process is completed within 6 hours.

We use the default CLIP vision encoder of LLaVA to encode all memory snapshots, frontier snapshots, egocentric views and image navigation targets. And the encoded vision features are further compressed to $12\times12$ (for image targets and egocentric views) and $3\times3$ (for memory snapshots and frontier snapshots) tokens in the training prompt.

During fine-tuning, we simultaneously optimize the model for exploration task and prefiltering task with cross-entropy loss. The loss weights for exploration and prefiltering are set to 1 and 0.3, respectively. The training goal of exploration is to correctly predict the ground truth choice of memory snapshot or frontier at each step. The training goal of prefiltering is to select the top 10 helpful objects that have been observed, based on the ground truth we collected earlier.

\section{Ablation Study}
\label{appendix_ablation}

\subsection{Ablation on Hyperparameter Choices}
We mainly evaluate on the number of egocentric observations at each step ($N$), the maximum distance an object should be included in the memory snapshot ($max\_dist$), and the number of prefiltered classes ($K$).

\begin{figure}[h]
    \centering
    \includegraphics[width=\linewidth]{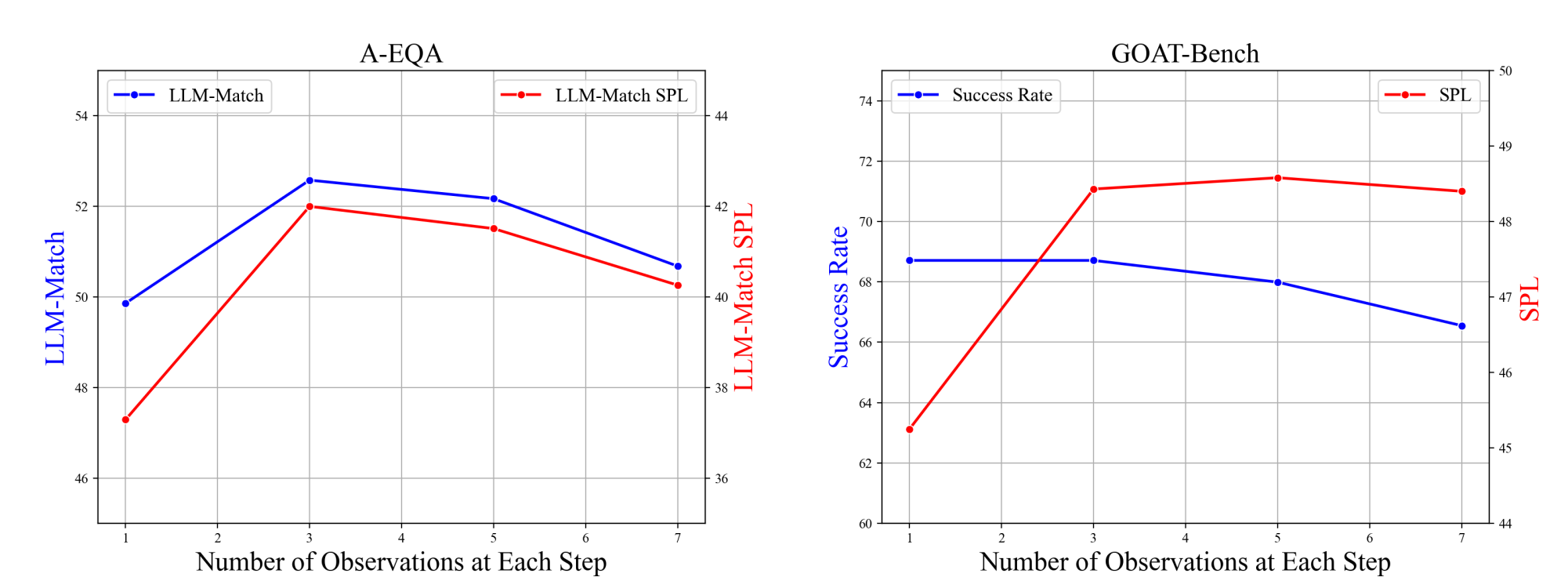}
    \caption{Ablation on the number of observation each step ($N$) for A-EQA and GOAT-Bench.}
    \label{fig:ablation_N}
\end{figure}

In Figure~\ref{fig:ablation_N}, we present the evaluation metrics for different choices of $N$ on both A-EQA and GOAT-Bench. We can observe that increasing the number of observations does not necessarily lead to better performance. This is mainly because the additional views often provide repeated and redundant information. Furthermore, as the number of frame candidates increases, a cluster of objects that would originally be assigned to one memory snapshots may instead be assigned to separate memory snapshots, resulting in confusion. Based on the results, we choose $N = 3$ for both datasets.

\begin{figure}[h]
    \centering
    \includegraphics[width=\linewidth]{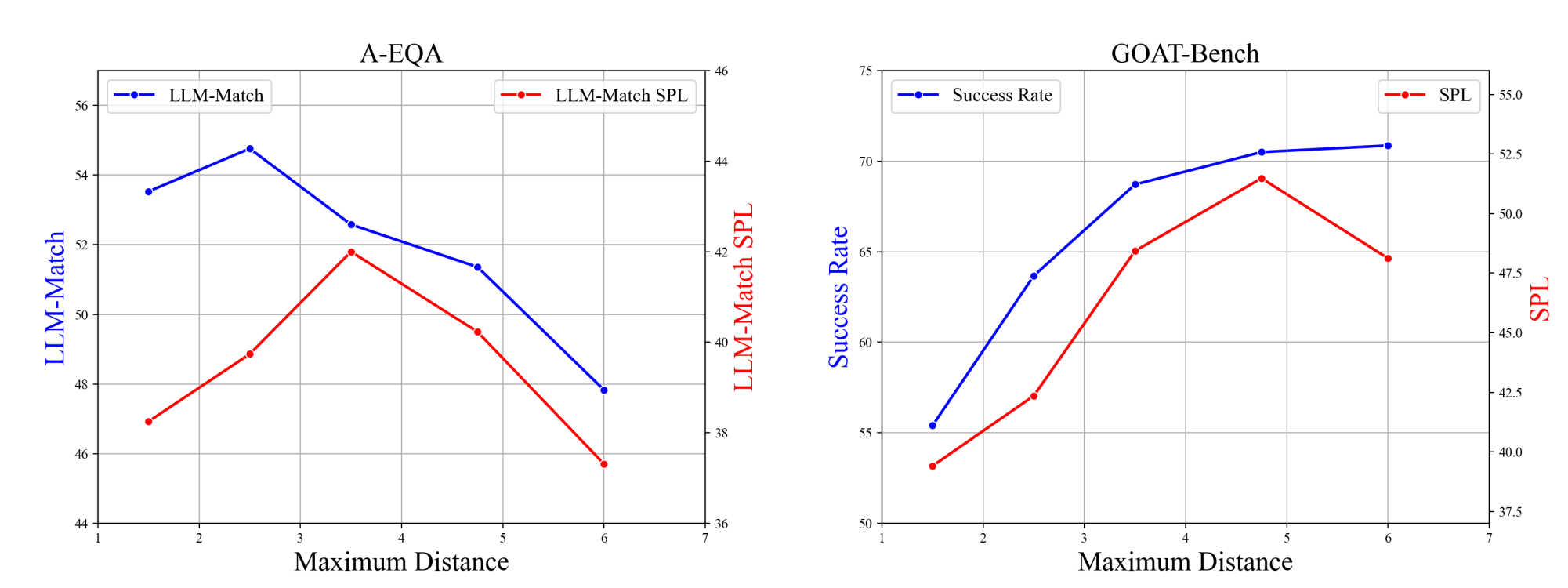}
    \caption{Ablation on the maximum distance for including an object to the scene graph ($max\_dist$) for A-EQA and GOAT-Bench.}
    \label{fig:ablation_min_dist}
\end{figure}

In Figure~\ref{fig:ablation_min_dist}, we present the evaluation metrics for different choices of $max\_dist$ on both A-EQA and GOAT-Bench, where we observe different tendencies across the two benchmarks. Evaluation metrics on GOAT-Bench generally improve with an increase in $max\_dist$, while metrics on A-EQA decline. This is because, under normal circumstances, a memory snapshot should only represent objects within a local area. Objects in more distant regions should either remain in unexplored areas or be captured by another memory snapshot that is closer to them. A large $max\_dist$ imposes a looser distance restriction, which can introduce disorder. However, in the navigation task of GOAT-Bench, the earlier the target object is added to the scene graph as a choice for the VLM, the faster the VLM can select it as the direct navigation target, resulting in faster arrival at the target objects. Balancing both accuracy and efficiency across the two benchmarks, we choose $max\_dist$ to be 3.5 meters.

\begin{figure}[h]
    \centering
    \includegraphics[width=\linewidth]{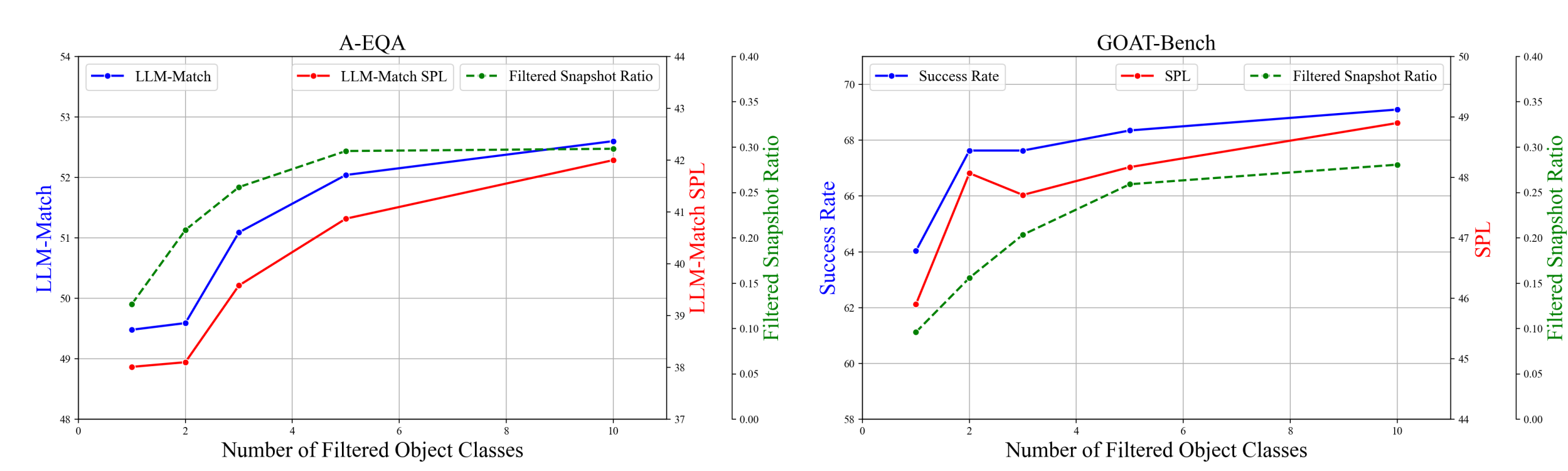}
    \caption{Ablation on the number of prefiltered classes ($K$) for A-EQA and GOAT-Bench.}
    \label{fig:ablation_K}
\end{figure}

In Figure~\ref{fig:ablation_K}, we present the evaluation metrics for different choices of $K$ on both A-EQA and GOAT-Bench. In addition to the metrics introduced in the experiment sections, we include the average ratio of the number of remaining memory snapshots after prefiltering to the total number of memory snapshots as a measure of the effectiveness and intensity of prefiltering. The results on both benchmarks align with our intuition: allowing more prefiltered classes leads to better performance. Moreover, even when $K=10$, on average only 3.26 and 4.66 memory snapshots are left after prefiltering for A-EQA and GOAT-Bench respectively, accounting for 29.8\% and 28.1\% of the total memory snapshots, and 8.2\% and 5.1\% of the total frame candidates. These statistics demonstrate the effectiveness of prefiltering as a memory retrieval mechanism, as well as 3D-Mem's compactness as a scene representation. Furthermore, we observe that the overall performance does not drop significantly when $K$ is small, highlighting the robustness of our framework.

\subsection{Ablation on Pipeline Components}
Ablation study on Prefiltering is infeasible because directly querying the VLM would exceed its context length. However, we conduct an ablation on Frontier Snapshots by always navigating to the nearest frontier when Memory Snapshots cannot provide the answer, rather than choosing a frontier via VLM. As shown in Table~\ref{tab:ablation_ft_snapshot} (SnapMem w/o FS), performance declines on both A-EQA and GOAT-Bench, though the drop is smaller on GOAT-Bench. This is likely due to the lifelong setting of GOAT-Bench, where the agent tends to rely on its memory once the scene is mostly explored. Additional experiments removing both Frontier Snapshots and memory maintenance (SnapMem w/o FS \& Mem) confirm this pattern.

\begin{table}[ht]
\renewcommand{\arraystretch}{0.5}
    \centering
    \resizebox{\linewidth}{!}{
    \begin{tabular}{lcc|cc}
        \toprule
        \multicolumn{1}{c}{} & 
        \multicolumn{2}{c}{\textbf{A-EQA}} & 
        \multicolumn{2}{c}{\textbf{GOAT-Bench}} \\
        \cmidrule(lr){2-3} \cmidrule(lr){4-5}
        \textbf{Method} & 
        \textbf{LLM-Match $\uparrow$} & 
        \textbf{LLM-Match SPL $\uparrow$} & 
        \textbf{Success Rate $\uparrow$} & 
        \textbf{SPL $\uparrow$} \\
        \midrule
        SnapMem w/o FS \& Mem & - & - & 57.2 & 33.2 \\
        SnapMem w/o FS           & 49.3 & 31.0 & 63.7 & 46.8 \\
        SnapMem                  & \textbf{52.6} & \textbf{42.0} & \textbf{69.1} & \textbf{48.9} \\
        \bottomrule
    \end{tabular}
    }
\caption{Ablation study of Frontier Snapshot on A-EQA and GOAT-Bench. FS denotes "Frontier Snapshots".}
\label{tab:ablation_ft_snapshot}
\end{table}

\section{Other Related Works}
While our work focuses on comparing to 3D scene representations, prior research on 2D scene representations, particularly in topological mapping for navigation, is notable. Methods like Topological Semantic Graph Memory~\cite{TSGM} and RoboHop~\cite{RoboHop} similarly represent environments as graphs using images and objects, where nodes correspond to images of navigable places and edges denote navigability.

Our proposed 3D-Mem first differs from these topological mapping methods in its focus and design. 3D-Mem focuses on capturing all salient objects in the scene by a minimum number of memory snapshots. Each memory snapshot is designed to capture the visual features of a cluster of objects in the nearby region, along with their spatial relationships and surrounding environment. Objects are uniquely assigned to one memory snapshot, making the representation informative, comprehensive, and compact—key qualities for leveraging vision-language models (VLMs) with limited context length to interpret and reason over visual data. In contrast, the images in topological map in~\cite{TSGM} are primarily designed to represent landmarks for navigation, without attempting to capture all informative aspects of the scene or focusing on visually representing all objects in a 3D environment. The images in the representation in~\cite{RoboHop}, though focused on object segments, are not compact, containing redundancy between consecutive frames, and the images still serve as navigation landmarks rather than visual representations of inter-object relationships.

In addition, 3D-Mem introduces the concept of frontier snapshots to explicitly model unexplored regions, allowing agents to make informed decisions about where to explore next to expand knowledge—an active exploration capability not addressed in previous 2D methods. Moreover, the structure of 3D-Mem enables the memory retrieval mechanism of Prefiltering, which manages the memory scalability and efficiency over extended operations, supporting lifelong learning that is absent in the aforementioned works. Lastly, like other 3D scene graphs, 3D-Mem stores the 3D information of the objects and snapshots. Based on this information, as in the practice of ConceptFusion~\citep{jatavallabhula2023conceptfusion}, a set of spatial relationship comparators can be called by LLMs as queries, \textit{e.g.}, querying the distance between A and B by calling ``howFar(A, B)''. This information is cannot be stored in those 2D representations.

\section{Complete Prompts for VLMs}
\label{appendix_full_prompts}
\begin{figure*}[h]
    \centering
    \includegraphics[width=\linewidth]{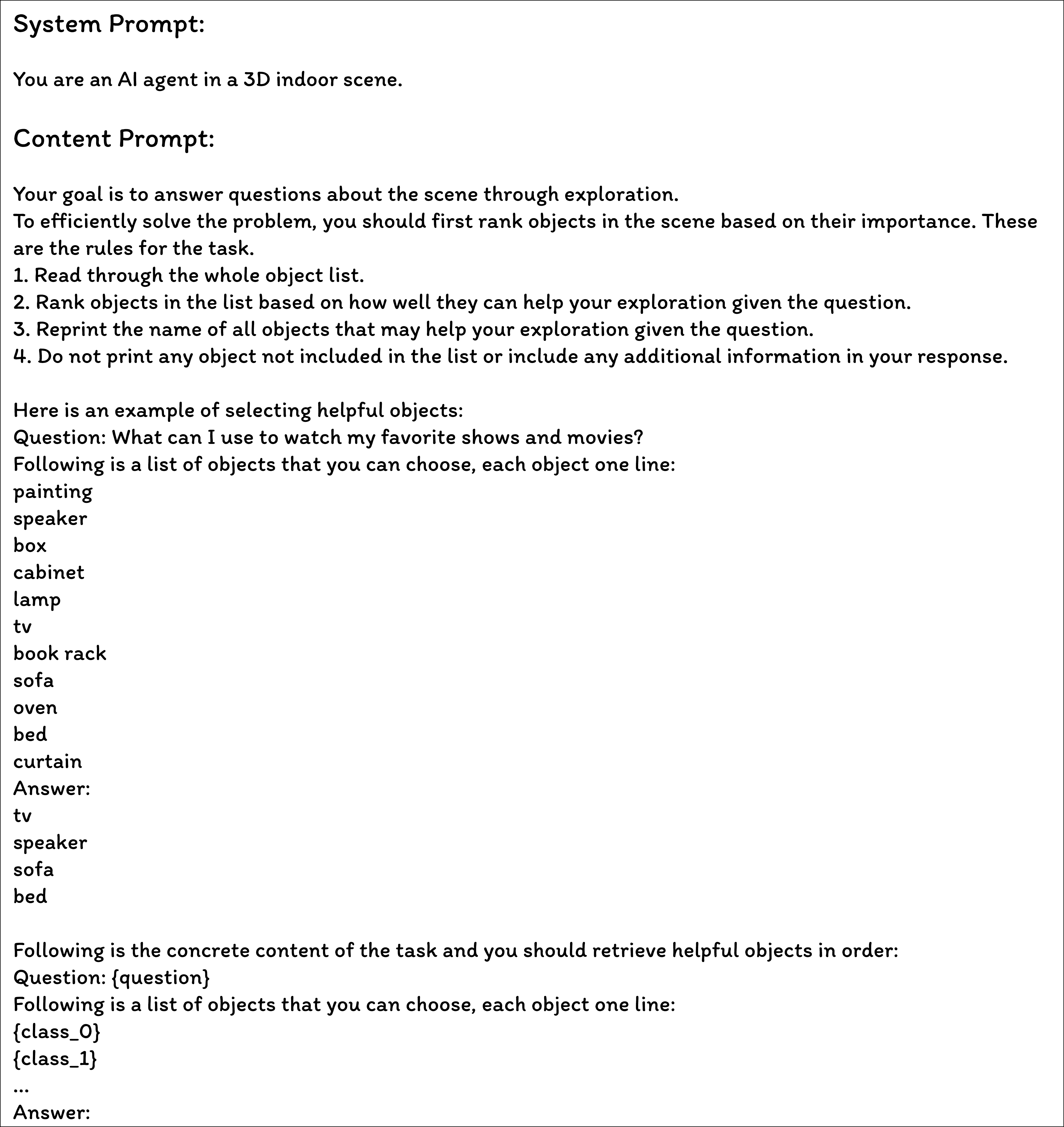}
    \caption{Prompt for prefiltering. The placeholders \{question\} and \{class\_$i$\} are replaced by the question and all existing classes in the scene graph, respectively.}
    \label{fig:appendix_prompt_prefiltering}
\end{figure*}

\begin{figure*}[h]
    \centering
    \includegraphics[width=\linewidth]{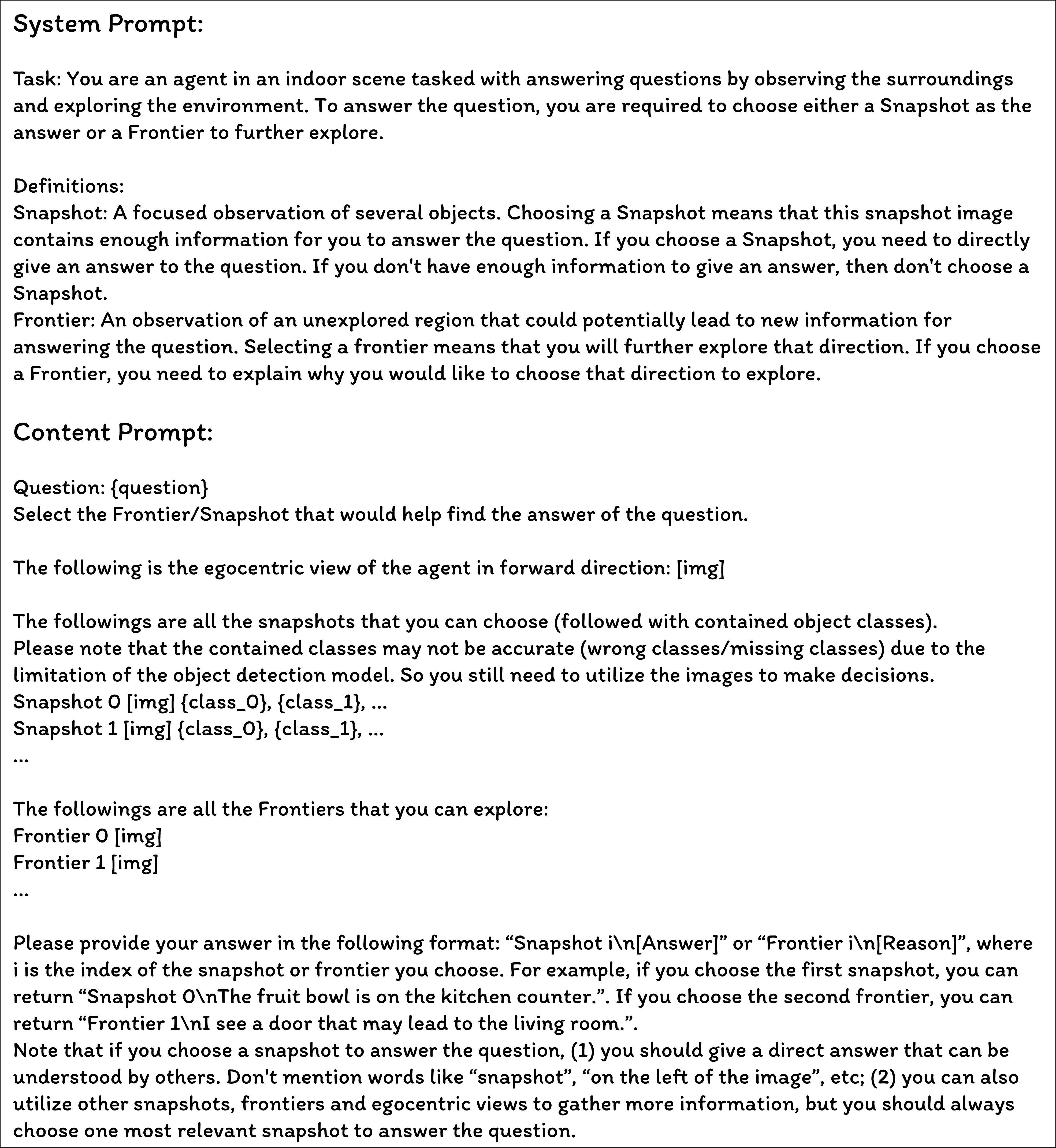}
    \caption{Prompt for embodied question answering. The placeholders \{question\} and \{class\_$i$\} are replaced by the question and the object classes contained in the corresponding memory snapshots, respectively. [img] are replaced by the egocentric views, memory snapshots or frontier snapshots.}
    \label{fig:appendix_prompt_aeqa}
\end{figure*}

\begin{figure*}
    \centering
    \includegraphics[width=\linewidth]{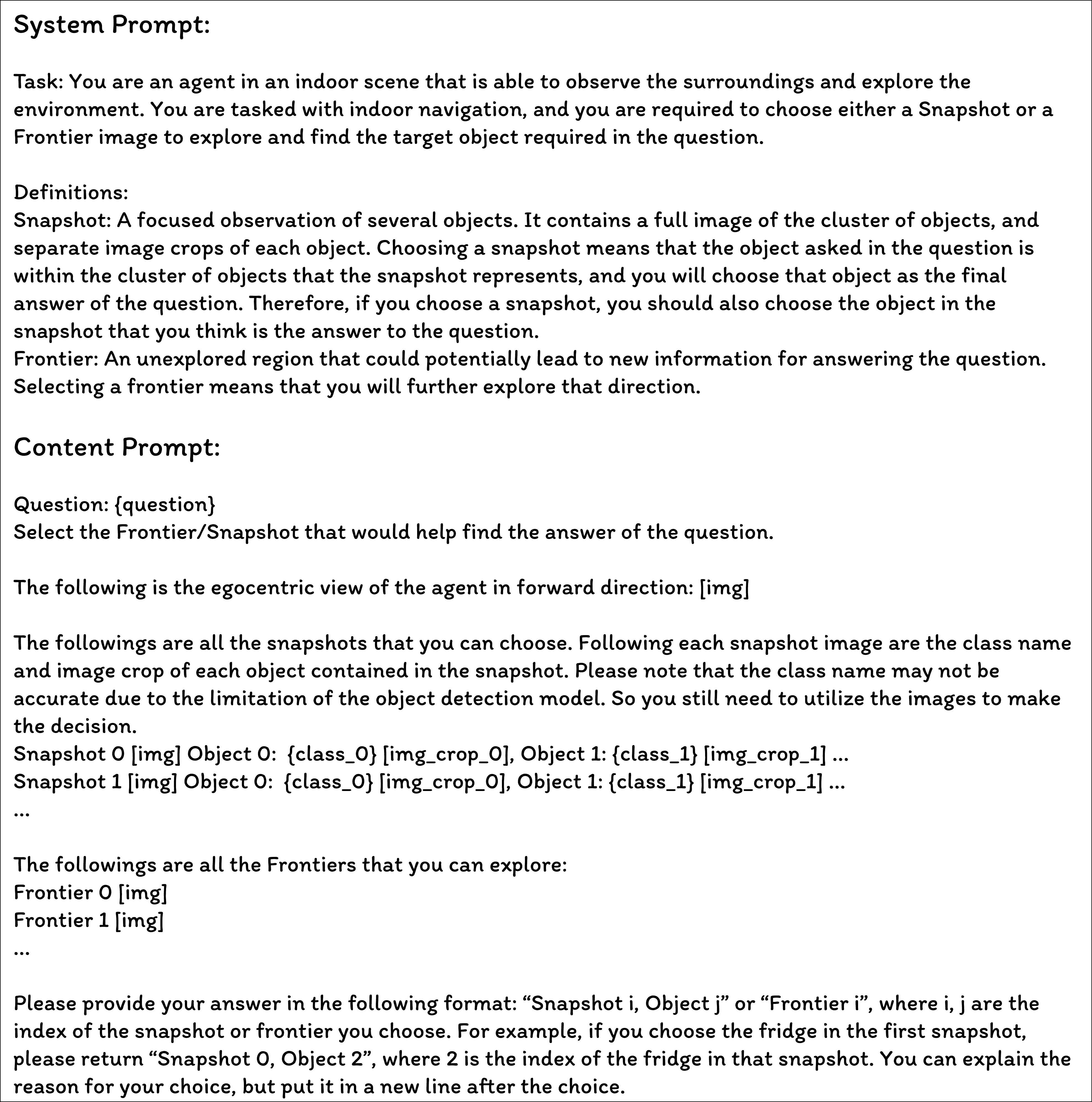}
    \caption{Prompt for GOAT-Bench dataset. The placeholders \{question\} and \{class\_$i$\} are replaced by the question and the object classes contained in the corresponding memory snapshots, respectively. [img] are replaced by the  egocentric views, memory snapshots or frontier snapshots, and [img\_crop\_$i$] are replaced by the corresponding object crops, which are directly cropped from the memory snapshots based on the detection bounding boxes.}
    \label{fig:appendix_prompt_goatbench}
\end{figure*}

We present the full prompt for prefiltering in Figure~\ref{fig:appendix_prompt_prefiltering}, the prompt for embodied question answering (A-EQA dataset) in Figure~\ref{fig:appendix_prompt_aeqa}, and the prompt for navigation (GOAT-Bench dataset) in Figure~\ref{fig:appendix_prompt_goatbench}.

\clearpage

\clearpage

\clearpage

{
    \small
    \clearpage
    \bibliographystyle{ieeenat_fullname}
    \bibliography{main}
}

\end{document}